# FlapKAD: A Simulation Dataset of Coupled Wing Kinematics and Aerodynamic Dynamics for Flapping-Wing Aerial Vehicles

**Haichuan Li**
haichuan.li@utu.fi
University of Turku

## Abstract

Experimental investigation and modeling of flapping-wing aerial vehicles are limited by the scarcity of large-scale records that temporally align wing kinematics, aerodynamic responses, and flight states. Existing datasets are often limited in scale and affected by measurement noise and temporal misalignment between rapidly varying wing motion and the associated dynamic response, particularly during high-frequency flapping. We introduce FlapKAD, an episode-structured simulation dataset comprising 2,000 rigid-wing flight episodes and 720,152 valid time steps, with bilateral wing kinematics, aerodynamic responses, and flight states recorded synchronously within a common clock. FlapKAD supports a unified bidirectional sequence-prediction benchmark constructed from the same temporally aligned episodes. The forward task predicts future vertical force coefficients and body vertical velocity from histories of realized flap and twist angles, whereas the inverse task reconstructs future flap- and twist-angle trajectories from the corresponding response histories. A benchmark of eight representative time-series architectures across multiple prediction horizons reveals direction- and horizon-dependent model behavior, systematically higher reconstruction errors for twist angle than for flap angle, and no consistent advantage from increased architectural complexity. FlapKAD provides a reproducible dataset and benchmark for studying coupled wing-kinematic, aerodynamic, and flight-state dynamics in flapping-wing aerial vehicles.

## Introduction

Flapping-wing aerial vehicles (FWAVs) offer a bio-inspired approach to small-scale flight, with potential advantages in maneuverability, compactness, and operation in confined environments. Unlike fixed-wing and rotary-wing platforms, FWAVs generate aerodynamic loads through periodic and strongly coupled variations in wing stroke, twist, local flow velocity, and angle of attack. The resulting aerodynamic response varies substantially within each wingbeat and depends on the preceding motion history. The relationship between wing kinematics and aerodynamic loading is therefore nonlinear, temporally structured, and difficult to represent using static or memoryless models(Sane and Dickinson 2002; Shyy et al. 2010). Accurately characterizing this relationship is important for aerodynamic analysis, surrogate modeling, and wing-motion reconstruction. A forward model maps a history of realized wing motion to future aerodynamic response, providing an efficient approximation of the underlying kinematic–aerodynamic dynamics. Conversely, an inverse model reconstructs future wing kinematics from aerodynamic and flight-state histories. The inverse direction is particularly challenging because the observed response can depend on multiple kinematic variables, their relative phase, the operating state, and the preceding trajectory. Sequence models provide a natural framework for both directions because they can represent temporal dependencies that are absent from instantaneous input–output formulations (Pereira et al. 2023; Sharvit, Karl, and Beatus 2025). Progress in this area is constrained by the scarcity of large-scale records that align rapidly varying wing kinematics with aerodynamic responses and flight states. Experimental acquisition becomes particularly difficult during rapid flapping, where measurement noise, calibration requirements, and temporal misalignment can obscure both instantaneous and history-dependent relationships. Existing forward and inverse studies are also commonly organized around separate datasets and task-specific protocols, limiting consistent comparison and reuse across the two directions.

To address these limitations, we introduce **FlapKAD**, a simulation dataset designed around episode-level temporal alignment. Rather than preparing separate representations for individual learning directions, FlapKAD synchronizes bilateral wing kinematics, wing-specific and aggregate aerodynamic responses, and flight states within each episode. Our dataset retains native variable-length trajectories for detailed analysis and provides compact fixed-length views for standardized benchmarking. Using this common representation, we formulate a bidirectional sequence-prediction benchmark in which the same episodes, channel definitions, partitions, and preprocessing procedures are used in both directions. The forward task predicts future vertical force coefficients and body vertical velocity from histories of realized flap and twist angles, whereas the inverse task reconstructs future realized flap- and twist-angle trajectories from the corresponding response histories. Eight representative time-series architectures are evaluated over short, intermediate, and long prediction horizons. The main contributions of this work are:

- **Comprehensive synchronized simulation dataset.** FlapKAD contains 2,000 rigid-wing flight episodes and 720,152 retained native simulator observations, syn-chronously pairing realized bilateral wing kinematics, aerodynamic responses, and flight states within a common episode structure.

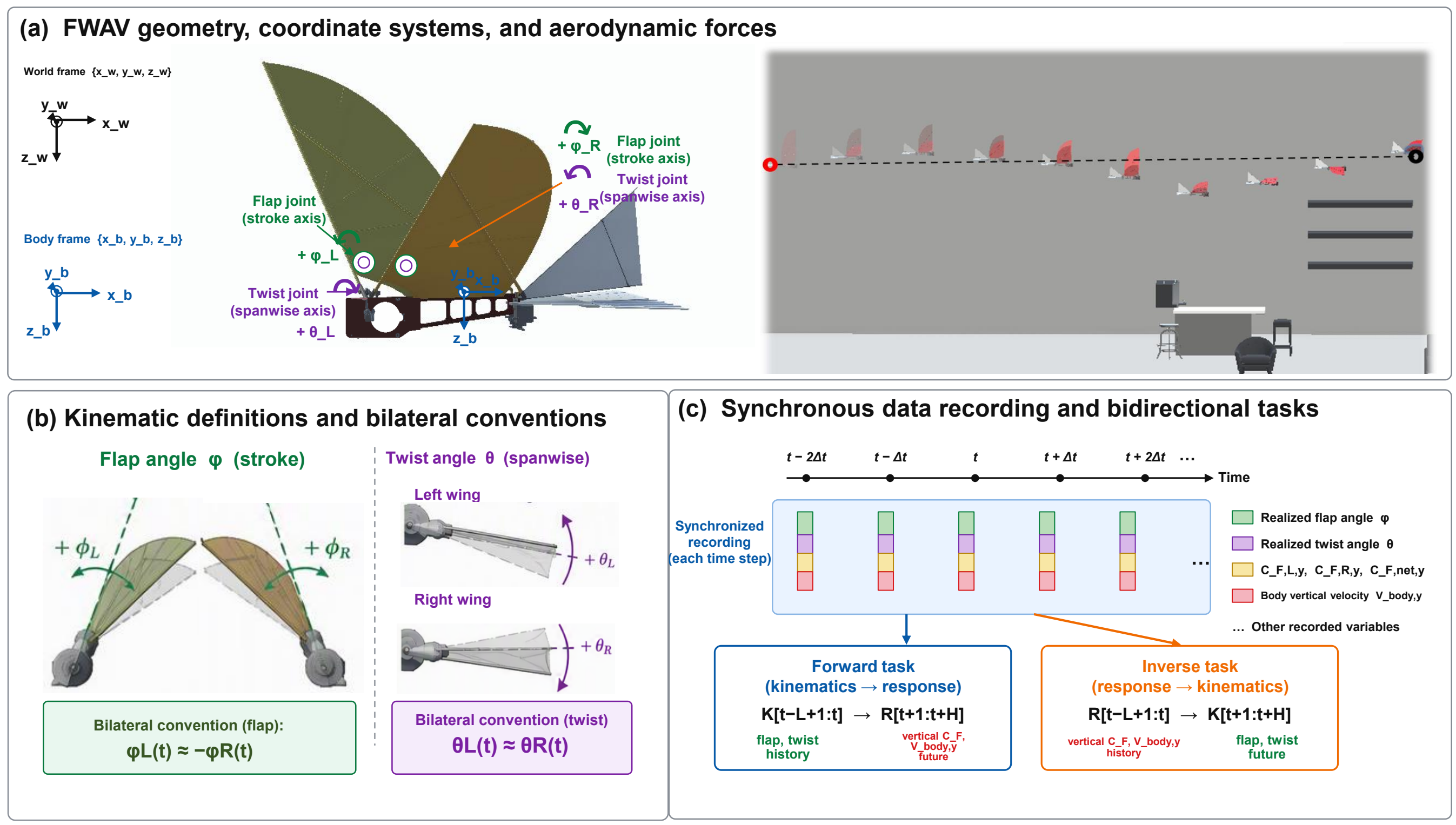

Figure 1: FWAV configuration, wing-kinematic conventions, and bidirectional sequence-prediction framework. (a) FWAV geometry, coordinate systems, wing joints, and experimental setup. (b) Bilateral conventions for flap and spanwise twist angles. (c) Synchronous recording and formulation of the forward and inverse prediction tasks.

- **Unified bidirectional benchmark.** Previous studies evaluated forward and inverse mappings separately, using task-specific datasets and experimental protocols that limit direct comparison across directions. FlapKAD provides a common episode-level interface(Fig.3), supports flexible temporal configurations, and eliminates the need to reconstruct separate task-specific datasets.
- **Multilevel release.** Native variable-length episodes are provided together with documented fixed-length views and validity masks, supporting both detailed analysis and standardized sequence learning.
- **Broad empirical evaluation.** Eight time-series architectures are evaluated in both task directions at $H \in \{1, 16, 32\}$ using common episode partitions, preprocessing, model selection, and metrics. The evaluation covers aggregate and channel-wise errors, parameter counts, and inference efficiency.

## Related Work

**Flapping-wing aerodynamic modeling.** Flapping-wing aerodynamic loading arises from strongly coupled translational, rotational, and wake-history effects. These mechanisms produce phase-dependent and transient responses that are difficult to represent using steady or memoryless formulations (Dickinson, Lehmann, and Sane 1999; Sane and Dickinson 2002; Sane 2003; Shyy et al. 2010). Data-driven and reduced-order approaches have been investigated for aerodynamic-load prediction and transient-response estimation (Bayiz and Cheng 2021b; Pereira et al. 2023; Lan et al. 2022), as well as for the surrogate-assisted optimization of flapping-wing kinematics (Corban, Bauerheim, and Jardin 2023). These studies demonstrate the value of learning temporal relationships between wing motion and aerodynamic response, but their task definitions and data organizations are generally developed for a particular modeling direction.

**Open flapping-wing datasets.** Open experimental resources have played an important role in learning-based flapping-wing research. Bayiz and Cheng released 548 synchronized trajectories containing wing motion, aerodynamic forces, and moments, and used these measurements to study history-dependent forward modeling (Bayiz and Cheng 2021b,a). Sharvit et al. subsequently released synchronized three-dimensional wing-motion and force measurements and demonstrated their use for inverse mapping from aerodynamic observations to wing kinematics (Sharvit, Karl, and Beatus 2025). These datasets provide valuable physical measurements and reproducible reference tasks. FlapKAD complements these resources with a larger and repeatable simulation dataset in which bilateral wing kinematics, aerodynamic responses, and flight states share a common temporal reference. Its episode-level organization enables forward and inverse benchmarks to be derived from the same underlying records and evaluated using common partitions, preprocessing procedures, and metrics.

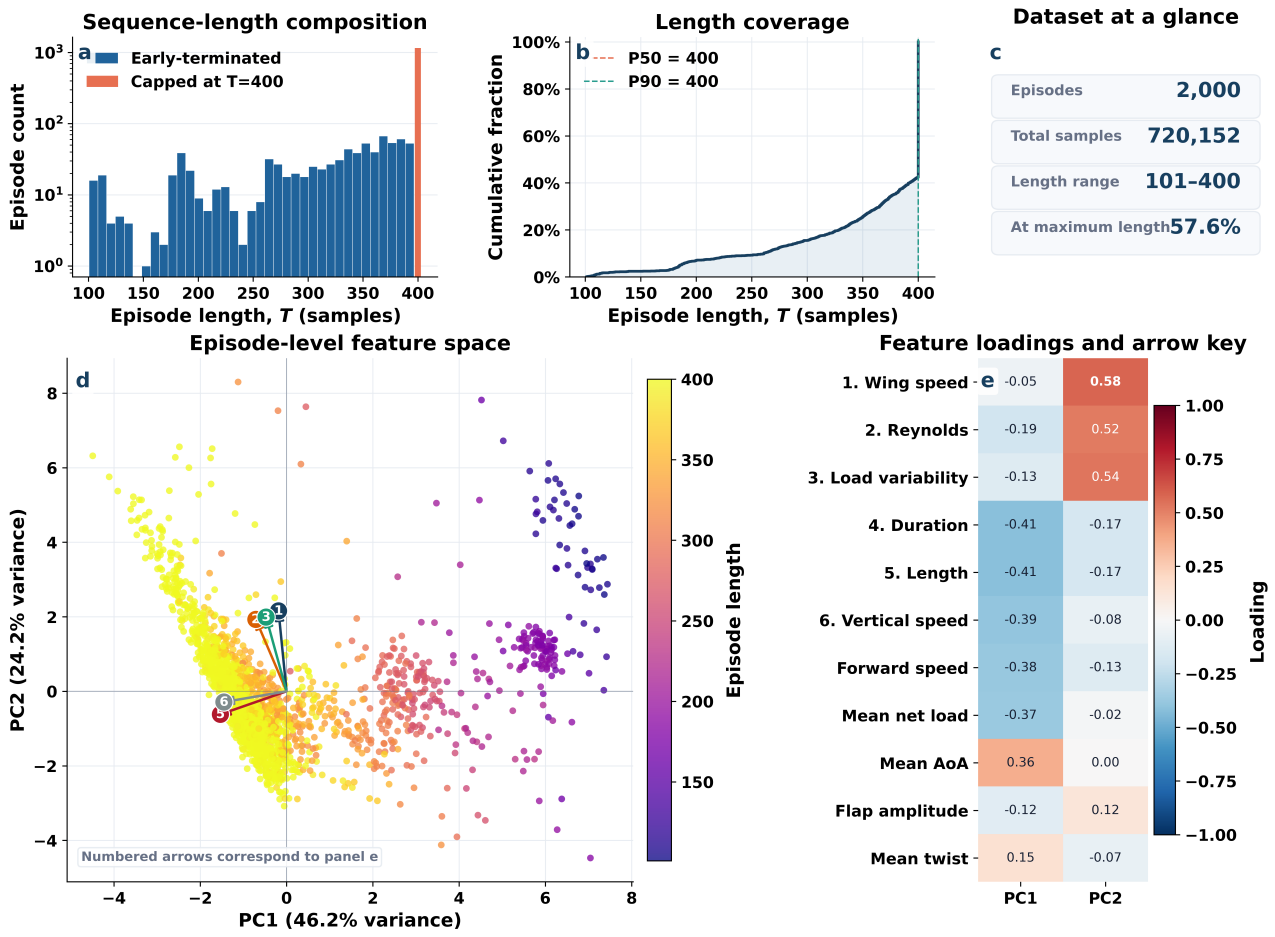

Figure 2: Dataset coverage and episode-level structure of FlapKAD. (a) Distribution of retained episode lengths, separating early-terminated episodes from episodes reaching the configured maximum length. (b) Empirical cumulative distribution of episode length. (c) Summary of dataset size, sample count, retained-length range. (d) Principal-component projection of episode-level summary features, colored by retained sequence length; numbered arrows indicate selected feature-loading directions. (e) Corresponding feature loadings for the first two principal components. The first two components explain 70.4% of the episode-level variance.

**Time-series forecasting architectures.** Our benchmark considers model families representing distinct temporal inductive biases. Recurrent networks propagate information through a learned hidden state, while temporal convolutional networks use causal and dilated convolutions to capture multiscale dependencies (Cho et al. 2014; Bai, Kolter, and Koltun 2018). DLinear and NLinear provide low-complexity linear baselines for evaluating whether direct temporal projection is sufficient (Zeng et al. 2023). Autoformer, PatchTST, iTransformer, and TimesNet represent decomposition-based, patch-based, variable-centered attention, and periodicity-aware approaches, respectively (Wu et al. 2021; Nie et al. 2023; Liu et al. 2024; Wu et al. 2023). Although these architectures are commonly evaluated on general-purpose forecasting datasets, their comparative behavior on flap-angle, twist-angle, aerodynamic-response, and flight-state trajectories remains less well characterized. FlapKAD provides a common bidirectional and multi-horizon setting for examining these families under consistent data and evaluation protocols.

# Methodology

## Dataset Collection and Physical Variables

FlapKAD was generated using a flapping-wing aerial vehicle simulator that integrates bilateral wing actua-tion, aerodynamic-force estimation, and synchronized data recording within a fixed-step loop. The aerodynamic contributions of the left and right wings are evaluated independently. For each wing, the relative-flow velocity at an aerodynamic sampling point is resolved with respect to the local chordwise, surface-normal, and spanwise directions. The local flow components determine the angle of attack, while the relative-flow magnitude determines the dynamic pressure used to evaluate lift and drag. The resulting force is applied to the vehicle rigid body at the prescribed aerodynamic force location. FlapKAD records the vertical components of the bilateral and aggregate aerodynamic forces along the simulator $y$ axis.

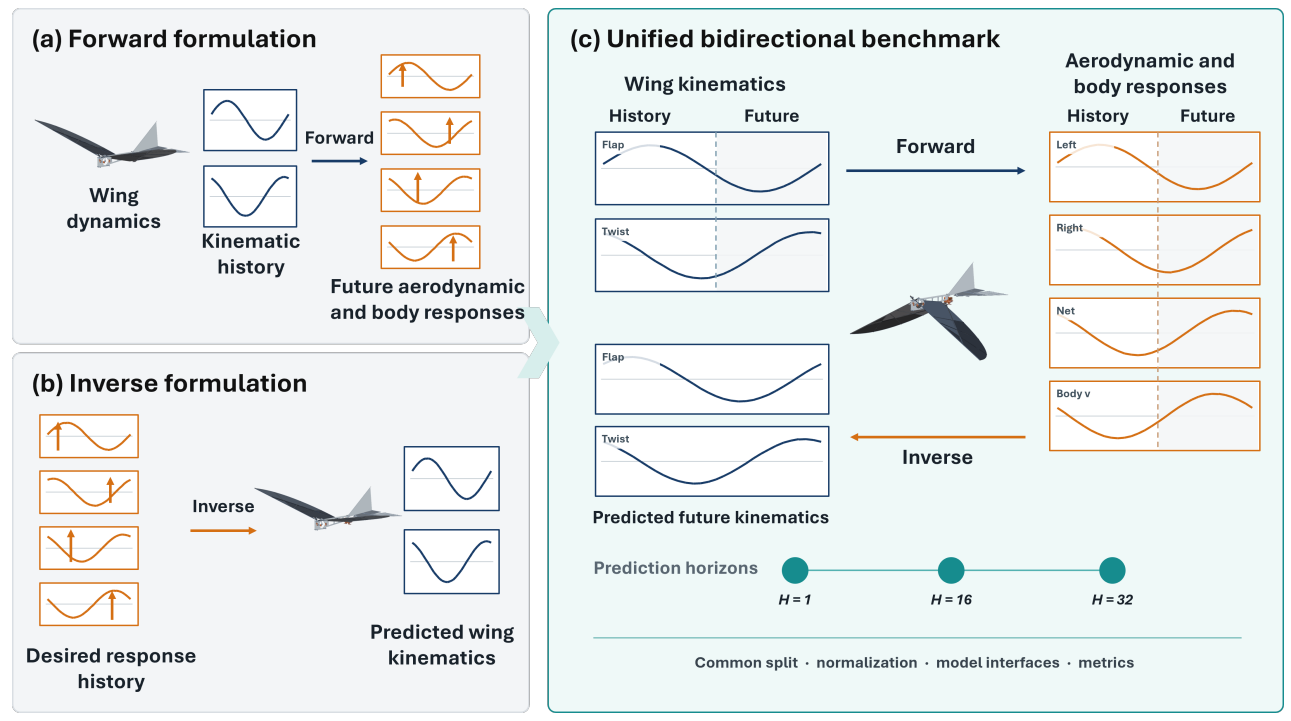

Figure 3: From unidirectional learning tasks to the unified FlapKAD benchmark. (a) Forward modelling predicts aerodynamic responses from wing-kinematic histories. (b) Inverse modelling recovers wing kinematics from response histories. Previous studies evaluated these directions separately. (c) FlapKAD unifies both tasks using shared episodes, data splits, normalization, model interfaces, and metrics.

Our benchmark kinematics comprise the realized left-wing flap and twist angles. The response variables comprise the left-wing, right-wing, and aggregate vertical force coefficients, together with the body vertical velocity. The aggregate vertical load is defined as

$$F_{\mathrm{net},y}(t) = F_{L,y}(t) + F_{R,y}(t). \tag{1}$$

The reference speed is the mean of the relative-flow speeds at the two wing sampling points:

$$U_{\mathrm{ref}}(t) = \frac{U_L(t) + U_R(t)}{2}, \tag{2}$$

where $U_L$ and $U_R$ are nonnegative speed magnitudes. The corresponding reference dynamic pressure and force-normalization denominator are

$$q_{\mathrm{ref}}(t) = \frac{1}{2}\rho U_{\mathrm{ref}}^2(t), \qquad D_F(t) = q_{\mathrm{ref}}(t)S, \tag{3}$$

where $\rho$ is the fluid density and $S$ is the common reference area. The vertical force coefficients are then

$$\begin{aligned} C_{F,L,y}(t) &= \frac{F_{L,y}(t)}{D_F(t)}, \qquad C_{F,R,y}(t) = \frac{F_{R,y}(t)}{D_F(t)}, \\ C_{F,\mathrm{net},y}(t) &= \frac{F_{\mathrm{net},y}(t)}{D_F(t)}. \end{aligned} \tag{4}$$

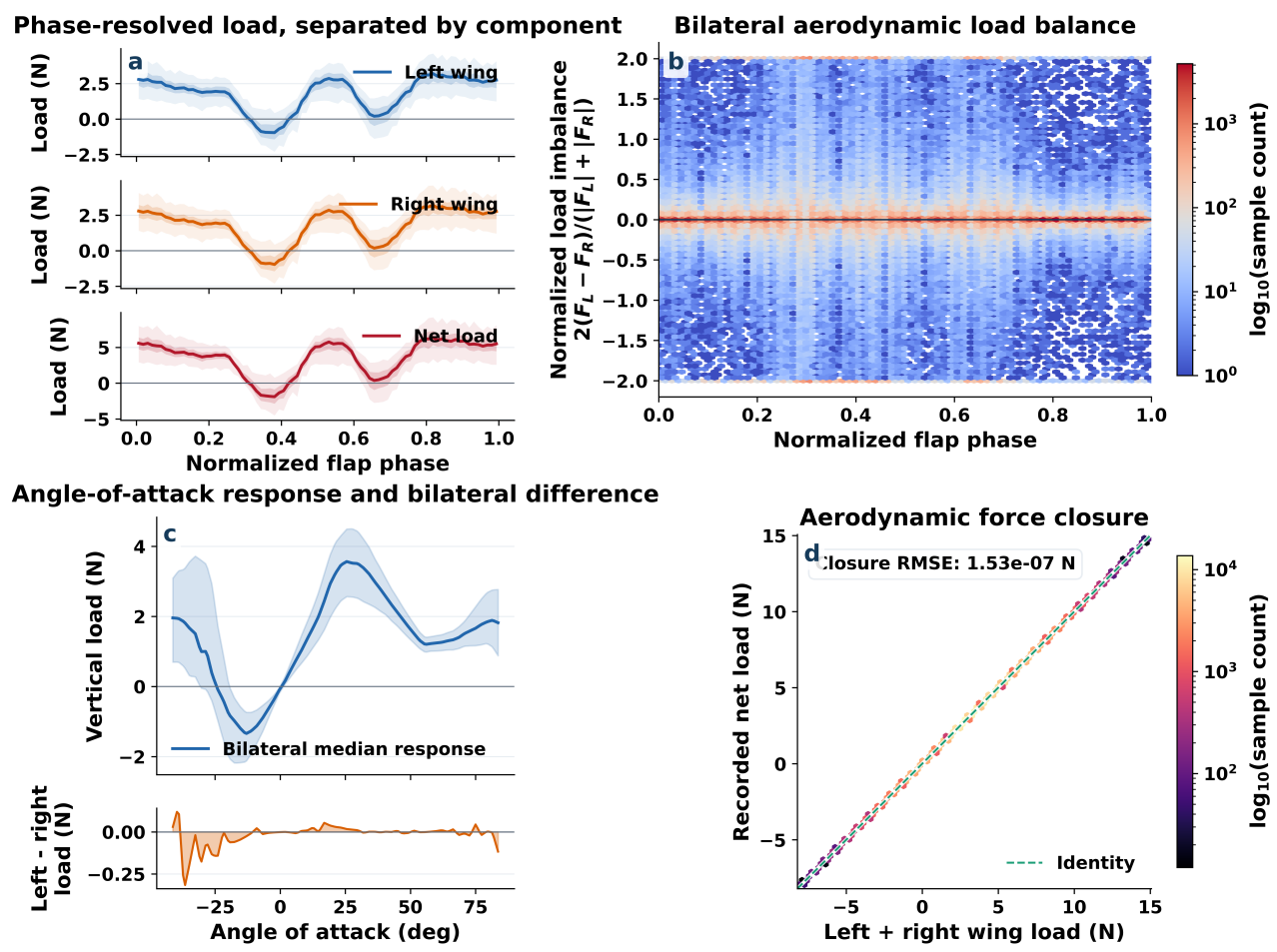

Figure 4: Aerodynamic response, bilateral load balance, and force consistency. (a) Phase-resolved left-wing, right-wing, and aggregate vertical aerodynamic loads; solid curves show median responses and shaded regions show episode-level envelopes. (b) Distribution of the normalized bilateral load imbalance $I_F$ over the flap cycle; the red curve indicates its phase-dependent central tendency. (c) Median vertical-load response as a function of angle of attack, together with the left-minus-right load difference. (d) Closure between the recorded aggregate vertical load and the sum of the left- and right-wing loads; the dashed line denotes exact equality.

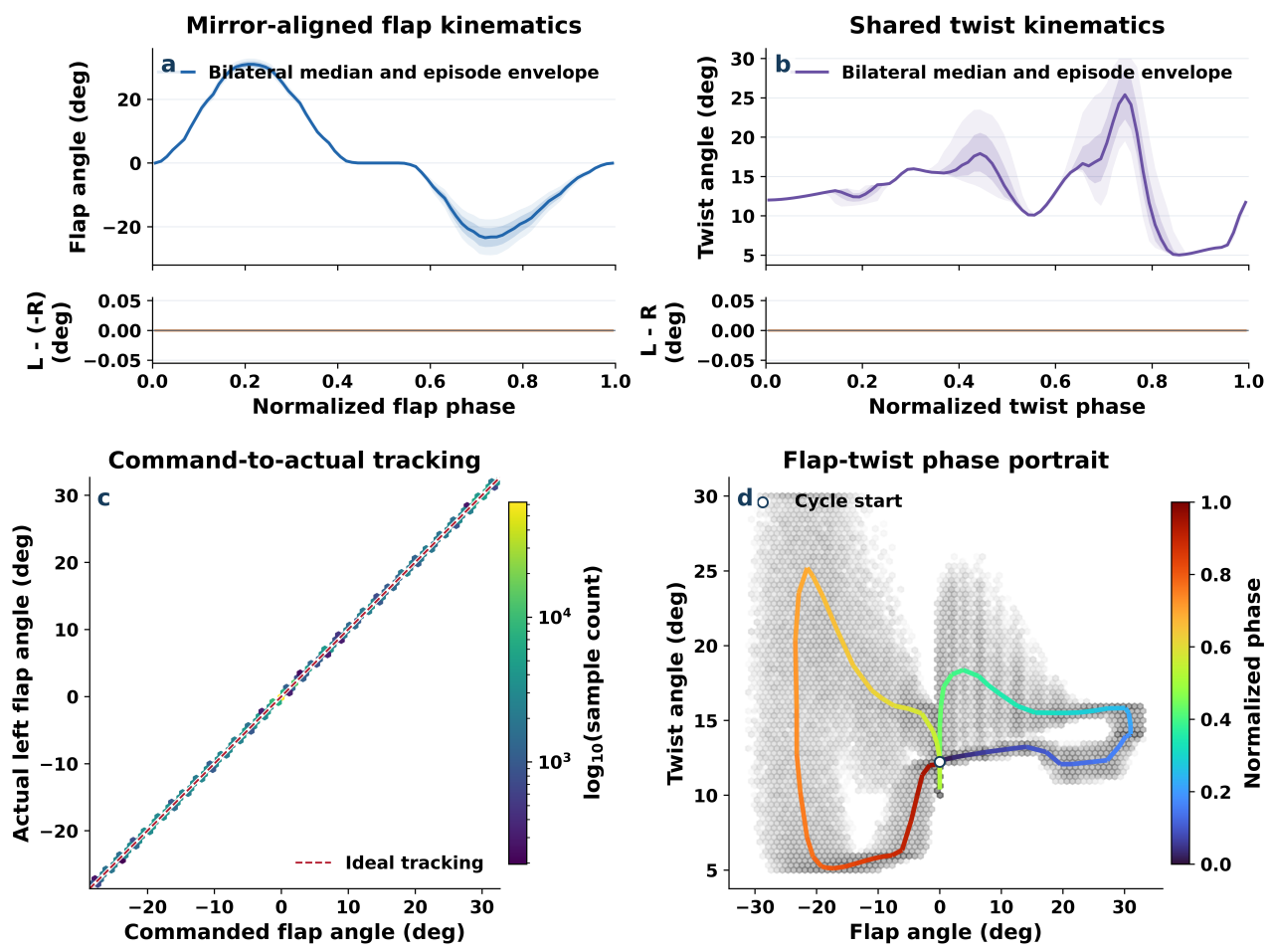

Figure 5: Bilateral kinematic consistency and flap-command tracking. (a) Mirror-aligned flap kinematics with median and episode envelope. (b) Shared twist-angle kinematics. (c) Commanded versus realized left-wing flap angles; the dashed line indicates ideal tracking. (d) Joint distribution of realized left- and right-wing flap angles, with median-cycle trajectories and episode-level density.

## Dataset Representation and Quality Control

Our data collection comprises 2,000 episodes, which boundaries are preserved during preprocessing, data splitting, and sequence construction, preventing temporal windows from crossing between independent episodes. The kinematic vector contains the realized left-wing flap and twist angles,

$$\mathbf{K}(t) = [\phi_L(t) \quad \theta_L(t)]^\top, \tag{5}$$

and the response vector contains the left-wing, right-wing, and aggregate vertical force coefficients together with the body vertical velocity,

$$\mathbf{R}(t) = [C_{F,L,y}(t) \quad C_{F,R,y}(t) \quad C_{F,\text{net},y}(t) \quad V_{\text{body},y}(t)]^\top. \tag{6}$$

The forward benchmark maps the two kinematic channels to the four response channels, whereas the inverse benchmark maps the same four response channels to the two kinematic channels. The representative left-wing kinematics are adopted because the realized bilateral trajectories follow the prescribed mirror convention for flap motion and the same-sign convention for twist motion, as further evaluated in the supplementary material.

Episodes are stored at a common length of 470 samples and accompanied by a binary validity mask. For episodes shorter than the storage length, the complete physical sequence is retained and the remaining positions are filled by repeating the final valid observation. The mask distinguishes valid observations from padded positions. It is used during sequence construction and the estimation of normalization statistics so that padded positions do not contribute to the benchmark samples or fitted statistics. Quality control is performed at the episode level before benchmark construction. Episodes containing non-finite core variables or failing the prescribed duration, wing-speed, or aggregate-load criteria are excluded. Each processed episode is associated with a quality record containing its original and retained lengths, retained duration, validity status, filtering reason, maximum bilateral wing speed, maximum absolute aggregate load, mean body velocity, and mean absolute angle of attack. Detailed channel definitions, units, conversion rules, filtering thresholds, episode-level quality statistics, and additional consistency analyses are provided in the supplementary material.

Figure 6 provides an overview of the dependence structure among the recorded kinematic, aerodynamic, and flight-state variables. The sign-aligned bilateral flap angles exhibit strong correlation, consistent with the mirrored wing convention. Structured associations are also observed between wing kinematics, vertical aerodynamic loading, and body motion, supporting their use in the paired forward and inverse prediction tasks. Because Pearson correlation captures only linear association, the matrix is interpreted as a descriptive summary rather than as evidence of causal or complete dynamical dependence.

## Unified Forward and Inverse Tasks

In contrast to task-specific datasets that encode a single predefined input–output mapping, FlapKAD provides a unified, task-agnostic representation of temporally aligned wing kinematics and aerodynamic responses. The same released episodes can therefore support both forward and inverse prediction, with the task direction determined by the selection of input and target variables rather than by separate dataset con-struction. Let $\mathbf{K}(t) \in \mathbb{R}^2$ denote the realized left-wing flap and twist angles, and let $\mathbf{R}(t) \in \mathbb{R}^4$ denote the three vertical force coefficients and body vertical velocity, as defined in Eqs. (5) and (6). For an input-history length $L$ and prediction horizon $H$, the input window contains the $L$ observations ending at time index $t$, and the target window contains the subsequent $H$ observations, as illustrated in Fig. 1. The forward and inverse tasks are formulated as

$$\begin{aligned} \widehat{\mathbf{R}}_{t+1:t+H} &= \mathcal{F}_{\boldsymbol{\theta}}\left(\mathbf{K}_{t-L+1:t}\right), \\ \widehat{\mathbf{K}}_{t+1:t+H} &= \mathcal{G}_{\boldsymbol{\psi}}\left(\mathbf{R}_{t-L+1:t}\right), \end{aligned} \tag{7}$$

where $\mathcal{F}_{\boldsymbol{\theta}}$ and $\mathcal{G}_{\boldsymbol{\psi}}$ denote the forward and inverse prediction models with learnable parameters $\boldsymbol{\theta}$ and $\boldsymbol{\psi}$, respectively.

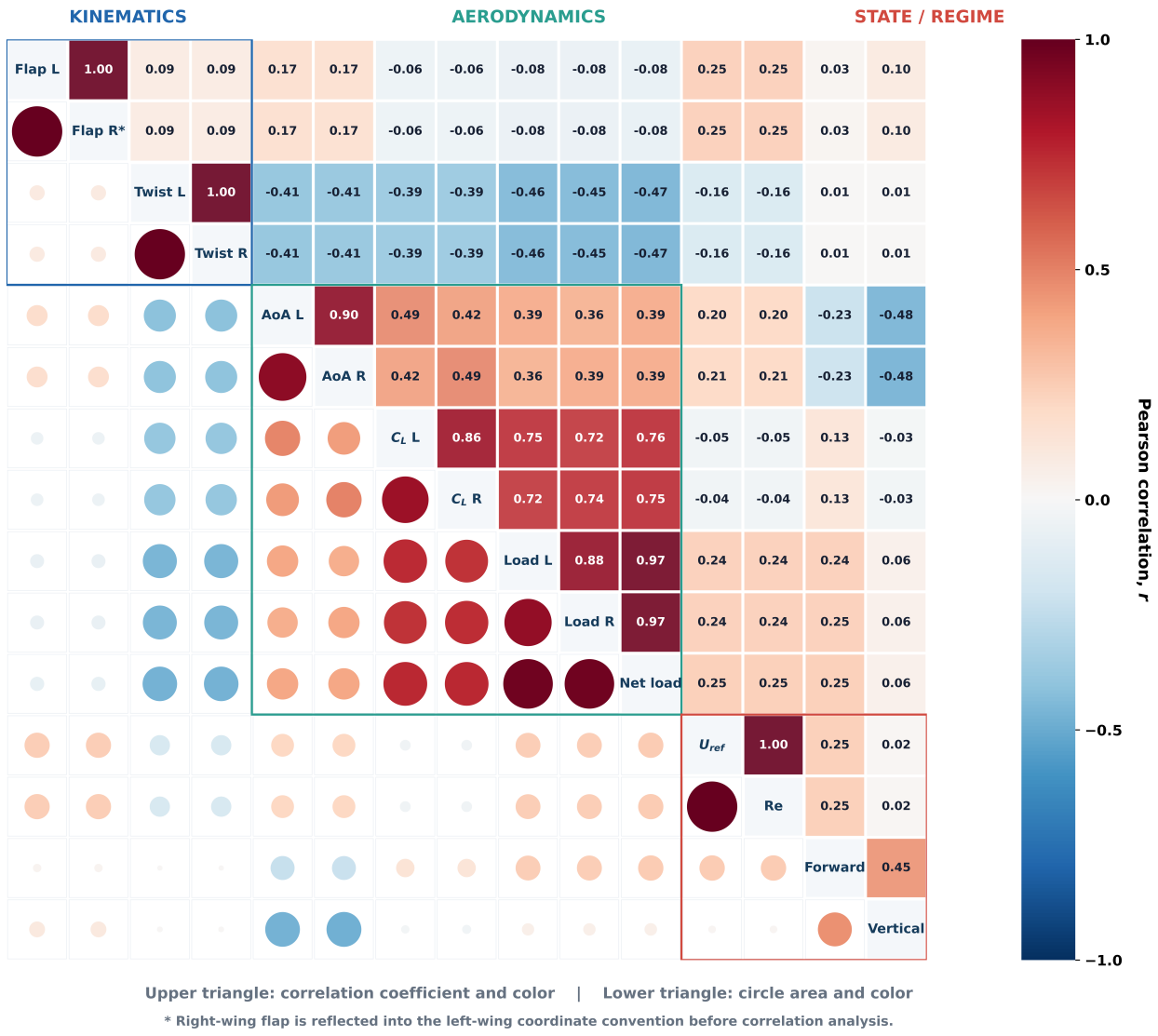

Figure 6: Pearson correlation structure among selected kinematic, aerodynamic, and flight-state variables. The upper triangle reports the correlation coefficients, while the lower triangle encodes correlation magnitude by circle area and correlation sign by color. Outlined blocks identify the kinematic, aerodynamic, and state or operating-condition variable groups. The right-wing flap angle is sign aligned with the left-wing convention before computing the correlations.

Forward task predicts future aerodynamic responses and body vertical velocity from a history of realized wing kinematics. Conversely, inverse task predicts future realized flap- and twist-angle trajectories from a history of the same response variables. FlapKAD preserves synchronized episode-level sequences rather than precomputed task-specific windows, allowing temporal configurations and task directions to be defined by the downstream data-loading pipeline.

# Experiments

## Experimental Scope

Our experiments evaluate FlapKAD from four complementary perspectives. First, we assess predictive performance in the forward and inverse task directions. Second, we examine how prediction accuracy and relative model performance vary with the prediction horizon. Third, we analyze variable-specific performance, including the individual aerodynamic-response variables in the forward task and the realized flap- and twist-angle trajectories in the inverse task. Finally, we examine the relationship between predictive accuracy and computational efficiency, with the complete efficiency results reported in the supplementary material.

Model performance is compared separately within each task direction and prediction horizon. Aggregate errors are not ranked directly across the forward and inverse tasks because the two directions have different target variables and output dimensions. In the result tables, the lowest value of each error metric is shown in bold, and the second-lowest value is underlined. When reported, the relative MSE improvement of model $a$ over reference model $b$ is defined as

$$\Delta_{a,b} = \frac{\mathrm{MSE}_b - \mathrm{MSE}_a}{\mathrm{MSE}_b} \times 100\%. \tag{8}$$

Relative improvements are computed for models evaluated under the same task direction, prediction horizon, episode split, target representation, and normalization setting.

## Benchmark Protocol

The acquired data preserve synchronized episode-level sequences rather than precomputed task-specific windows. This representation allows users to flexibly select input and target variables and define task directions, history lengths, prediction horizons, and window strides for downstream applications. For the reported benchmark, the data-loading pipeline dynamically constructs samples using an input-history length of $L = 256$, prediction horizons $H \in \{1, 16, 32\}$, and a stride of one. This episode-level representation allows the same released data to support different temporal configurations and task directions.

Our dataset indexing, an input–target pair beginning at valid index $s$ in episode $i$ is defined as:

$$\mathbf{X}_{i,s} = \mathbf{Z}_{i,s:s+L-1}, \qquad \mathbf{Y}_{i,s} = \mathbf{W}_{i,s+L:s+L+H-1}, \tag{9}$$

where $(\mathbf{Z}, \mathbf{W}) = (\mathbf{K}, \mathbf{R})$ for forward prediction and $(\mathbf{Z}, \mathbf{W}) = (\mathbf{R}, \mathbf{K})$ for inverse prediction. Let $T_i$ denote the mask-defined valid length of episode $i$. A starting index is admissible only if

$$s + L + H \leq T_i. \tag{10}$$

The target window therefore contains the $H$ observations immediately following the $L$-step input history. Every input–target pair lies entirely within the valid portion of a single episode, thus no sample crosses an episode boundary or includes padded positions. The 2,000 episodes are partitioned before window construction into 1,600 training, 200 validation, and 200 test episodes. A fixed split manifest is reused across task directions, prediction horizons, and model architectures. All windows derived from a given episode remain in the same partition, preventing overlapping segments of the same trajectory from appearing in both model development and final test evaluation. Each selected input and target channel is standardized independently using statistics estimated exclusively from valid observations in the training episodes:

$$\widetilde{x}_{t,j} = \frac{x_{t,j} - \mu_j^x}{\sigma_j^x}, \qquad \widetilde{y}_{t,k} = \frac{y_{t,k} - \mu_k^y}{\sigma_k^y}. \tag{11}$$

Here, $\mu_j^x$ and $\sigma_j^x$ denote the training-derived mean and standard deviation of input channel $j$, and $\mu_k^y$ and $\sigma_k^y$ are defined analogously for target channel $k$. Padded positions and all validation and test observations are excluded when estimating these statistics. The fitted transformations are then applied unchanged during validation and testing.

### Training and Evaluation

Our benchmark evaluates eight representative architectures through a common forecasting interface. Where required, a task-specific output adapter maps the native model output to

$$\widehat{\mathbf{Y}} \in \mathbb{R}^{B \times H \times D_y}, \tag{12}$$

where $B$ is the batch size, $H$ is the prediction horizon, and $D_y$ is the number of target channels.

All models follow a common optimization and model-selection protocol. The best checkpoint for each model, task direction, and prediction horizon is selected according to validation MSE and subsequently evaluated on the held-out test partition. Predictive performance is evaluated in normalized target space using mean squared error, mean absolute error, and the 95th percentile of the absolute error:

$$\mathrm{MSE} = \frac{1}{N}\sum_{n=1}^{N}(\widehat{y}_n - y_n)^2, \qquad \mathrm{MAE} = \frac{1}{N}\sum_{n=1}^{N}|\widehat{y}_n - y_n|. \tag{13}$$

$$E_{95} = Q_{0.95}\left(\{|\widehat{y}_n - y_n|\}_{n=1}^{N}\right), \tag{14}$$

where $N$ is the total number of scalar target values and $Q_{0.95}$ denotes the empirical 95th percentile. Channel-wise MSE and MAE are additionally reported for the individual target variables. Complete optimization settings, architecture-specific configurations, metric-aggregation details, and inference-efficiency protocols are provided in the supplementary material.

## Results

### Overall Benchmark Performance

Tables 1 and 2 summarize the normalized test errors for all 48 model–task–horizon configurations. The forward benchmark exhibits a relatively stable architecture ranking across prediction horizons, whereas inverse performance is substantially more horizon dependent. No single architecture attains the lowest MSE across all evaluated settings.

### Forward Prediction

Table 1 shows that TCN achieves the lowest forward MSE at all three prediction horizons, although its margins over the closest competing models are small. At $H = 1$, TCN obtains an MSE of $0.734$, compared with $0.741$ for GRU. TCN also achieves the lowest MAE, whereas GRU obtains the lowest $E_{95}$, indicating that the model with the lowest average error does not necessarily minimize the upper tail of the absolute-error distribution. At $H = 16$, TCN and GRU are nearly tied by MSE, both rounding to $0.703$, although TCN ranks first using the unrounded values. At $H = 32$, TCN achieves the lowest MSE, MAE, and $E_{95}$, with values of $0.649$, $0.584$, and $1.580$, respectively. PatchTST ranks second across all three metrics, while iTransformer, TimesNet, and GRU form a closely grouped set behind the two leading models. DLinear, NLinear, and Autoformer remain less accurate than the leading nonlinear models across the evaluated horizons. Several nonlinear models also exhibit lower aggregate errors at larger values of $H$. This trend does not imply that longer-horizon prediction is intrinsically easier, because changing $H$ also changes the admissible-window population, evaluated target positions, output dimension, and empirical target distribution. Cross-horizon differences are therefore interpreted as benchmark-specific behavior rather than as a controlled measure of forecasting difficulty.

### Inverse Prediction

Table 2 shows that inverse-model performance depends strongly on the prediction horizon. At $H = 1$, GRU achieves the lowest MSE, MAE, and $E_{95}$, with values of $0.052$, $0.127$, and $0.473$, respectively. At $H = 16$, TCN ranks first across all three metrics, followed by PatchTST. At $H = 32$, PatchTST becomes the leading model, attaining an MSE of $0.078$, an MAE of $0.169$, and an $E_{95}$ of $0.566$. TCN ranks second by MSE and $E_{95}$, whereas iTransformer obtains the second-lowest MAE. GRU exhibits the clearest horizon-dependent degradation, with its MSE increasing from $0.052$ at $H = 1$ to $0.100$ at $H = 16$ and $0.234$ at $H = 32$. By contrast, TimesNet remains comparatively stable across the three horizons. Autoformer outperforms the two linear baselines but remains substantially less accurate than the leading nonlinear models. These results show that strong one-step inverse performance does not necessarily extend to multi-step trajectory prediction. To visualize the performance margins between architectures, the relative MSE gap of model $m$ within task direction $\tau$ and prediction horizon $H$ is defined as

$$G_{m,H}^{(\tau)} = \frac{\mathrm{MSE}_{m,H}^{(\tau)} - \min_{m'} \mathrm{MSE}_{m',H}^{(\tau)}}{\min_{m'} \mathrm{MSE}_{m',H}^{(\tau)}} \times 100\%. \tag{15}$$

A value of zero identifies the best-performing architecture in the corresponding task–horizon setting.

Figure 7 provides a complementary view of the benchmark rankings. The leading forward models remain relatively close across horizons, whereas inverse performance is substantially more architecture and horizon dependent. In particular, the inverse winner changes from GRU at $H = 1$ to TCN at $H = 16$ and PatchTST at $H = 32$. Model selection for inverse prediction should therefore be conducted at the horizon relevant to the intended application rather than inferred from one-step performance alone.

### Flap- and Twist-Angle Prediction

Aggregate inverse metrics conceal a consistent difference between the two kinematic targets. Across all evaluated models and prediction horizons, the normalized twist-angle MSE is higher than the corresponding flap-angle MSE. For the best aggregate model at each horizon, the flap- and twist-angle MSE values are $0.011$ and $0.092$ for GRU at $H = 1$, $0.020$ and $0.137$ for TCN at $H = 16$, and $0.024$ and $0.132$ for PatchTST at $H = 32$. The persistence of this difference across model families suggests that the larger twist-angle error is not specific to a single architecture. Possible contributing factors include differences in target dynamics, observability from the selected response variables, and aerodynamic sensitivity to flap and twist motion; however, the present benchmark does not isolate the underlying cause. These results demonstrate the importance of channel-wise evaluation: a low aggregate inverse error can coexist with accurate flap-angle prediction and substantially larger twist-angle error. Aggregate and channel-wise metrics should therefore be considered together when evaluating inverse models.

| Model | $H = 1$ | | | $H = 16$ | | | $H = 32$ | | |
|---|---|---|---|---|---|---|---|---|---|
| | MSE | MAE | $E_{95}$ | MSE | MAE | $E_{95}$ | MSE | MAE | $E_{95}$ |
| DLinear | 0.850 | 0.683 | 1.786 | 0.861 | 0.694 | 1.770 | 0.808 | 0.679 | 1.694 |
| NLinear | 0.852 | 0.684 | 1.787 | 0.862 | 0.695 | 1.770 | 0.808 | 0.679 | 1.694 |
| GRU | <u>0.741</u> | <u>0.614</u> | **1.740** | <u>0.703</u> | <u>0.607</u> | 1.673 | 0.668 | 0.598 | 1.594 |
| TCN | **0.734** | **0.611** | <u>1.745</u> | **0.703** | **0.606** | **1.666** | **0.649** | **0.584** | **1.580** |
| PatchTST | 0.754 | 0.619 | 1.753 | 0.718 | 0.610 | <u>1.670</u> | <u>0.656</u> | <u>0.585</u> | <u>1.582</u> |
| TimesNet | 0.764 | 0.625 | 1.761 | 0.724 | 0.607 | 1.715 | 0.666 | 0.586 | 1.601 |
| iTransformer | 0.765 | 0.623 | 1.784 | 0.719 | 0.610 | 1.707 | 0.662 | 0.588 | 1.606 |
| Autoformer | 0.796 | 0.651 | 1.810 | 0.787 | 0.642 | 1.773 | 0.775 | 0.654 | 1.704 |

Table 1: Normalized test errors for forward prediction. Lower values indicate better performance. Within each prediction horizon, bold and underlined entries denote the best and second-best results for each metric, respectively. Rankings are determined using the unrounded values.

| Model | $H = 1$ | | | $H = 16$ | | | $H = 32$ | | |
|---|---|---|---|---|---|---|---|---|---|
| | MSE | MAE | $E_{95}$ | MSE | MAE | $E_{95}$ | MSE | MAE | $E_{95}$ |
| DLinear | 0.628 | 0.585 | 1.579 | 0.688 | 0.617 | 1.633 | 0.739 | 0.641 | 1.676 |
| NLinear | 0.633 | 0.586 | 1.596 | 0.693 | 0.618 | 1.642 | 0.742 | 0.643 | 1.677 |
| GRU | **0.052** | **0.127** | **0.473** | 0.100 | 0.192 | 0.687 | 0.234 | 0.314 | 1.053 |
| TCN | <u>0.078</u> | <u>0.152</u> | <u>0.586</u> | **0.079** | **0.166** | **0.592** | <u>0.094</u> | 0.184 | <u>0.653</u> |
| PatchTST | 0.087 | 0.176 | 0.602 | <u>0.086</u> | <u>0.174</u> | <u>0.613</u> | **0.078** | **0.169** | **0.566** |
| TimesNet | 0.111 | 0.186 | 0.717 | 0.115 | 0.189 | 0.726 | 0.113 | 0.187 | 0.716 |
| iTransformer | 0.120 | 0.191 | 0.729 | 0.101 | 0.180 | 0.680 | 0.101 | <u>0.181</u> | 0.677 |
| Autoformer | 0.339 | 0.396 | 1.265 | 0.433 | 0.454 | 1.418 | 0.576 | 0.543 | 1.628 |

Table 2: Normalized test errors for inverse prediction. Lower values indicate better performance. Within each prediction horizon, bold and underlined entries denote the best and second-best results for each metric, respectively. Rankings are determined using the unrounded values.

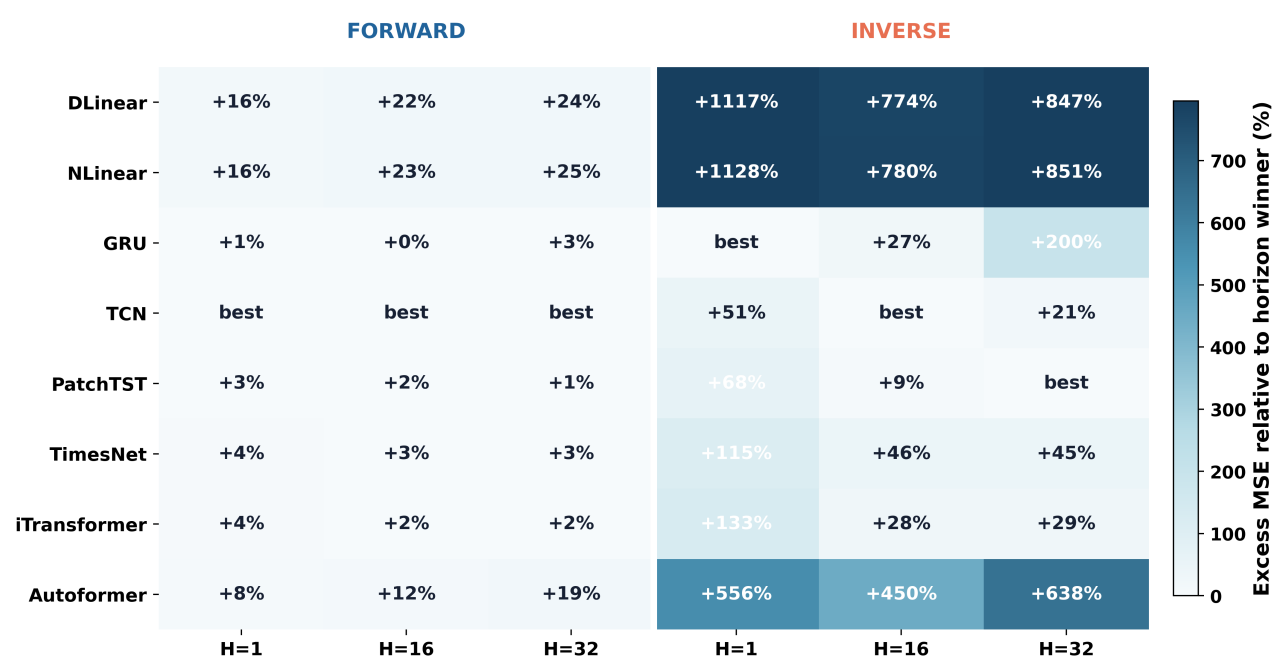

Figure 7: Relative MSE gap from the best-performing architecture within each task direction and prediction horizon. Zero identifies the setting-specific winner, while positive values indicate the percentage by which a model's MSE exceeds the minimum MSE. Forward and inverse values are normalized independently.

## Conclusion

This work introduced FlapKAD, an episode-level simulation dataset and benchmark for learning the coupled relationship between wing kinematics and aerodynamic responses in flapping-wing aerial vehicles. By preserving synchronized kinematic, aerodynamic, and flight-state sequences rather than predefined task-specific samples, FlapKAD supports flexible construction of forward and inverse prediction tasks, channel selections, and temporal configurations. The benchmark demonstrates the importance of evaluating multiple prediction horizons and reporting both aggregate and channel-wise performance. FlapKAD provides a reproducible foundation for developing, screening, and comparing sequence-learning methods before physical FWAV testing, where sensor payload or apparatus intrusion, calibration, measurement noise, and temporal synchronization complicate the precise acquisition of rapidly varying kinematic and aerodynamic quantities.

## Appendix

# Simulation and Aerodynamic Validation

## Simulation and Data Acquisition

FlapKAD was generated using a rigid-wing flapping-wing aerial vehicle simulation implemented in Unity. The simulation integrates rigid-body dynamics, bilateral wing actuation, aerodynamic-force estimation, flight-control components, episode management, and synchronized data logging within a common fixed-step loop.

Aerodynamic forces are evaluated independently for the left and right wings. For each wing, the simulator estimates the velocity of a moving aerodynamic sampling point relative to the ambient flow and resolves the resulting velocity along the local chordwise, surface-normal, and spanwise directions. The local flow components determine the angle of attack, while the relative-flow magnitude determines the dynamic pressure used to evaluate lift and drag. The resulting aerodynamic force is applied to the vehicle rigid body at the prescribed force location.

Each episode begins from a reset state and includes a settling interval before data acquisition. During this interval, rigid-body motion and aerodynamic-force application are temporarily suspended to prevent reset transients from being recorded as physical observations. Logging starts after the settling procedure has completed and the controller and aerodynamic components have been restored.

Episode conditions are varied through the configured initialization, gait, operating-condition, and controller settings. An episode terminates when its maximum duration is reached or when predefined safety or validity conditions are violated. Episode boundaries and termination information are retained explicitly. At each recorded time step, the simulator synchronously records wing kinematics, aerodynamic quantities, and vehicle states. The public benchmark release retains the variables required by the forward and inverse prediction tasks, while the broader simulator records are used during dataset construction and consistency analysis.

## Aerodynamic Variable Definitions

FlapKAD uses the average-wing-speed convention to define the aerodynamic reference speed. Let $U_L(t)$ and $U_R(t)$ denote the nonnegative relative-flow speed magnitudes at the left- and right-wing aerodynamic sampling points, respectively. The reference speed is

$$U_{\mathrm{ref}}(t) = \frac{U_L(t) + U_R(t)}{2}. \tag{16}$$

The corresponding reference dynamic pressure is

$$q_{\mathrm{ref}}(t) = \frac{1}{2}\rho U_{\mathrm{ref}}^2(t), \tag{17}$$

where $\rho$ denotes the fluid density.

The vertical aerodynamic components are expressed along the Unity world-frame $y$-axis. The aggregate vertical load is defined as

$$F_{\mathrm{net},y}(t) = F_{L,y}(t) + F_{R,y}(t), \tag{18}$$

where $F_{L,y}$ and $F_{R,y}$ are the left- and right-wing vertical aerodynamic loads.

A common force-normalization denominator is used for the bilateral and aggregate force coefficients:

$$D_F(t) = q_{\mathrm{ref}}(t)S, \tag{19}$$

where $S$ is the reference area used by the simulator. The corresponding vertical force coefficients are

$$C_{F,L,y}(t) = \frac{F_{L,y}(t)}{D_F(t)}, \qquad C_{F,R,y}(t) = \frac{F_{R,y}(t)}{D_F(t)}, \qquad C_{F,\mathrm{net},y}(t) = \frac{F_{\mathrm{net},y}(t)}{D_F(t)}. \tag{20}$$

The recorded Reynolds number and simulator-defined span-to-chord ratio are

$$Re(t) = \frac{U_{\mathrm{ref}}(t)c}{\nu}, \qquad AR_{\mathrm{sim}} = \frac{R}{c}, \tag{21}$$

where $c$, $R$, and $\nu$ denote the reference chord, the wing span used by the simulator convention, and the kinematic viscosity, respectively. The notation $AR_{\mathrm{sim}}$ distinguishes the recorded span-to-chord ratio from alternative aspect-ratio definitions based on total span and planform area.

## Aerodynamic Consistency Checks

Bilateral aerodynamic balance is evaluated using the normalized load imbalance

$$I_F(t) = \frac{2\left[F_{L,y}(t) - F_{R,y}(t)\right]}{|F_{L,y}(t)| + |F_{R,y}(t)| + \epsilon_F}, \tag{22}$$

where $\epsilon_F > 0$ is a small regularization constant that prevents division by zero. Values near zero indicate balanced bilateral loading, whereas the sign of $I_F$ identifies which wing produces the larger vertical load.

The phase-resolved left- and right-wing loads exhibit similar patterns over the flap cycle. Their differences are reflected in the distribution of $I_F$, which is concentrated near zero while retaining the variation associated with asymmetric operating conditions across episodes.

The angle-of-attack analysis reveals a nonlinear relationship between local angle of attack and vertical aerodynamic loading. The observed dispersion is consistent with the additional dependence of the instantaneous load on relative-flow speed, dynamic pressure, flap phase, and other time-varying aerodynamic conditions.

Internal force consistency is evaluated by comparing the aggregate load recorded by the simulator with the sum of the two wing-specific contributions. The force-closure residual is

$$r_F(t) = F_{\mathrm{main},y}(t) - \left[F_{L,y}(t) + F_{R,y}(t)\right], \tag{23}$$

where $F_{\mathrm{main},y}$ denotes the recorded aggregate channel. The root-mean-square closure error is

$$\mathrm{RMSE}_F = \sqrt{\frac{1}{N}\sum_{i=1}^{N} r_F^2(t_i)}, \tag{24}$$

where $N$ is the total number of valid evaluated time steps. The measured closure error is

$$\mathrm{RMSE}_F = 1.53 \times 10^{-7}\,\mathrm{N}, \tag{25}$$

indicating numerical consistency between the recorded aggregate load and the sum of the bilateral wing contributions.

## External Validation of Our Synthetic Dataset Against Real-World Experimental Data

### Cross-Dataset Evaluation Protocol

To examine whether the temporal learning patterns in our synthetic dataset are also observed in real-world flapping-wing measurements, we repeat the same forecasting benchmark on the independent real-world experimental dataset.

The synthetic and real-world datasets use the same converted tensor representation, episode-level window construction, training-derived normalization procedure, model implementations, prediction horizons, validation-based checkpoint selection, and normalized-space evaluation metrics. Both benchmarks use a 256-sample input history and direct prediction horizons

$$H \in \{1, 16, 32\}.$$

The comparison is restricted to the shared channel subspace. The inverse task maps four aerodynamic channels to two kinematic channels,

$$\mathbf{F}^{\text{shared}}_{s:s+256} \longrightarrow \mathbf{U}^{\text{shared}}_{s+256:s+256+H}, \tag{26}$$

whereas the forward task maps the same two kinematic channels to the four aerodynamic channels,

$$\mathbf{U}^{\text{shared}}_{s:s+256} \longrightarrow \mathbf{F}^{\text{shared}}_{s+256:s+256+H}. \tag{27}$$

More specifically, the inverse benchmark uses aerodynamic channels $[0, 1, 2, 4]$ as inputs and kinematic channels $[0, 2]$ as targets. The forward benchmark reverses this mapping. The channel order is fixed across the two datasets.

Each dataset is partitioned independently at the episode level before window construction. Each dataset is also standardized using only its own training episodes. Consequently, normalized error magnitudes describe task difficulty relative to each dataset's training distribution and should not be interpreted as direct equality of physical error between simulation and experiment. We evaluate the seven model families available in both benchmarks: DLinear, NLinear, GRU, TCN, PatchTST, TimesNet, and iTransformer. Autoformer is included in the synthetic benchmark but excluded from the cross-dataset ranking analysis because the corresponding real-world runs were not performed. All results in this section use model seed 42 to preserve a directly matched architecture-level comparison.

### Inverse Prediction on Our Synthetic Dataset

Table 4 reports inverse-prediction performance on our synthetic flapping-wing dataset. GRU obtains the lowest error at $H = 1$, whereas TCN and PatchTST obtain the best MSE at $H = 16$ and $H = 32$, respectively. DLinear and NLinear remain substantially less accurate than the temporal neural models. The change in the leading architecture with prediction horizon indicates that long-horizon inverse prediction in the synthetic dataset favours models capable of representing longer-range or patch-level structure.

### Inverse Prediction on Real-World Data

Table 3 reports the corresponding results on the independent real-world experimental dataset under the same shared channel definitions and training protocol.

GRU obtains the lowest MSE, MAE, and P95 at all three horizons in the real-world benchmark. TCN ranks second at $H = 1$, whereas TimesNet is the second-best model at $H = 16$ and $H = 32$. The linear baselines remain near the bottom of the ranking, as in the synthetic benchmark.

The leading model groups agree strongly at $H = 1$, where GRU and TCN rank first and second in both datasets. The rankings diverge at longer inverse horizons. In particular, PatchTST improves with horizon in the synthetic benchmark but remains substantially less accurate on the real-world data.

### Forward Prediction on Our Synthetic Dataset

TCN obtains the lowest MSE at all three horizons on the synthetic forward task. GRU remains in the leading group at $H = 1$ and $H = 16$, while PatchTST becomes the second-best model at $H = 32$. DLinear and NLinear consistently obtain the highest errors.

### Forward Prediction on Real-World Experimental Data

TCN and GRU form the leading model group in both datasets. TCN obtains the lowest real-world error at $H = 1$, while GRU obtains the lowest error at $H = 16$ and $H = 32$. The linear baselines again occupy the lowest-performing group.

### Overall Agreement Between Our Synthetic and Real-World Benchmarks

To quantify architecture-level agreement, we compare the MSE rankings obtained on our synthetic dataset with those obtained on the independent real-world experimental dataset. The analysis includes the seven model families evaluated in both datasets: DLinear, NLinear, GRU, TCN, PatchTST, TimesNet, and iTransformer. Figure 8 summarizes the comparison using three complementary indicators: Spearman rank correlation, overlap between the two leading architectures, and the identities of those leading models. Exact numerical results are reported in Table 7.

The six task settings form three agreement groups. First, inverse prediction at $H = 1$ and forward prediction at $H \in \{1, 16\}$ show strong agreement, with rank correlations between $0.857$ and $0.893$ and complete Top-2 overlap. In these settings, GRU and TCN form the leading architecture group in both datasets. Second, forward prediction at $H = 32$ shows moderate agreement, with $\rho = 0.523$ and one shared Top-2 architecture. TCN remains in the leading group, but GRU rises from fifth on our synthetic dataset to first on the real-world data. Third, inverse prediction at $H = 16$ and $H = 32$ shows weaker agreement and no common Top-2 architectures. The rank correlations decline to $0.536$ and $0.286$, respectively. The loss of agreement is therefore concentrated in medium- and long-horizon inverse prediction rather than distributed uniformly across all evaluated tasks.

| Model | $H = 1$ | | | $H = 16$ | | | $H = 32$ | | |
|---|---|---|---|---|---|---|---|---|---|
| | MSE | MAE | P95 | MSE | MAE | P95 | MSE | MAE | P95 |
| DLinear | 0.489 | 0.558 | 1.303 | 0.473 | 0.539 | 1.368 | 0.492 | 0.556 | 1.364 |
| NLinear | 0.478 | 0.520 | 1.415 | 0.477 | 0.542 | 1.377 | 0.482 | 0.539 | 1.400 |
| GRU | **0.031** | **0.122** | **0.369** | **0.031** | **0.123** | **0.381** | **0.045** | **0.136** | **0.422** |
| TCN | 0.049 | 0.157 | 0.446 | 0.060 | 0.175 | 0.498 | 0.099 | 0.224 | 0.645 |
| PatchTST | 0.444 | 0.521 | 1.376 | 0.454 | 0.530 | 1.368 | 0.453 | 0.515 | 1.403 |
| TimesNet | 0.055 | 0.172 | 0.502 | 0.050 | 0.164 | 0.438 | 0.057 | 0.172 | 0.504 |
| iTransformer | 0.098 | 0.234 | 0.631 | 0.104 | 0.240 | 0.676 | 0.111 | 0.240 | 0.676 |

Table 3: Inverse-prediction performance on the independent real-world experimental flapping-wing dataset using model seed 42. The task uses the same shared aerodynamic and kinematic channels as the synthetic benchmark. Errors are computed in normalized target space. Bold and underlined entries denote the best and second-best values within each prediction horizon.

| Model | $H = 1$ | | | $H = 16$ | | | $H = 32$ | | |
|---|---|---|---|---|---|---|---|---|---|
| | MSE | MAE | P95 | MSE | MAE | P95 | MSE | MAE | P95 |
| DLinear | 0.628 | 0.585 | 1.579 | 0.688 | 0.617 | 1.633 | 0.739 | 0.641 | 1.676 |
| NLinear | 0.633 | 0.586 | 1.596 | 0.693 | 0.618 | 1.642 | 0.742 | 0.643 | 1.677 |
| GRU | **0.052** | **0.127** | **0.473** | 0.100 | 0.192 | 0.687 | 0.234 | 0.314 | 1.053 |
| TCN | 0.078 | 0.152 | 0.586 | **0.079** | **0.166** | **0.592** | 0.094 | 0.184 | 0.653 |
| PatchTST | 0.087 | 0.176 | 0.602 | 0.086 | 0.174 | 0.613 | **0.078** | **0.169** | **0.566** |
| TimesNet | 0.111 | 0.186 | 0.717 | 0.115 | 0.189 | 0.726 | 0.113 | 0.187 | 0.716 |
| iTransformer | 0.120 | 0.191 | 0.729 | 0.101 | 0.180 | 0.680 | 0.101 | 0.181 | 0.677 |

Table 4: Inverse-prediction performance on our synthetic flapping-wing dataset using model seed 42. Errors are computed in normalized target space. Bold and underlined entries denote the best and second-best values within each prediction horizon. Rankings are determined from unrounded results.

## Architecture-Ranking Transfer and Horizon-Dependent Gap

Figure 9 shows how the MSE rank of each architecture changes from our synthetic benchmark to the independent real-world benchmark. The visualization distinguishes architectures whose relative performance transfers across datasets from those whose ranking is domain dependent.

For inverse prediction at $H = 1$, GRU and TCN retain the first and second positions in both datasets. The corresponding rank correlation is $0.857$, and both Top-2 architectures are shared. This agreement indicates that our synthetic dataset reproduces the principal architecture-level pattern of short-horizon inverse prediction observed in real-world measurements.

The ranking differences increase with inverse-prediction horizon. At $H = 16$, TCN and PatchTST form the leading pair on our synthetic dataset, whereas GRU and TimesNet lead on the real-world data. At $H = 32$, PatchTST moves from first on our synthetic dataset to fifth on the real-world data, while GRU moves from fifth to first. TimesNet similarly improves from fourth to second. These changes explain the lower inverse rank correlations at $H = 16$ and $H = 32$.

The divergence is also apparent in horizon-dependent error growth. On the real-world inverse task, the TCN MSE at $H = 32$ is approximately twice its value at $H = 1$, while TimesNet remains comparatively stable. On our synthetic data, PatchTST and TCN become the leading long-horizon inverse models, whereas GRU exhibits considerably greater degradation. The two datasets therefore differ in which architecture best represents long-range inverse structure.

Forward prediction shows stronger cross-dataset agreement. TCN and GRU occupy the two leading positions in both datasets at $H = 1$ and $H = 16$. At $H = 32$, TCN remains in the leading group, while GRU rises from fifth on our synthetic dataset to first on the real-world data. Despite this change, recurrent and convolutional architectures remain among the strongest forward models in both domains, whereas DLinear and NLinear consistently occupy the lowest-performing group.

## Interpretation and Scope of the External Validation

Figure 10 provides the paired architecture ranks underlying the correlation statistics. Points on the dashed diagonal retain the same rank across the two datasets, whereas displacement from the diagonal indicates a change in relative model performance.

The short-horizon inverse panel and the forward panels at $H = 1$ and $H = 16$ remain comparatively concentrated around the diagonal. The larger off-diagonal displacements at inverse $H = 16$ and $H = 32$ are primarily associated with GRU, PatchTST, and TimesNet, consistent with the rank changes shown in Figure 9. The disagreement therefore reflects specific architecture-dependent changes rather than a uniform reversal of all model rankings.

| Model | $H = 1$ | | | $H = 16$ | | | $H = 32$ | | |
|---|---|---|---|---|---|---|---|---|---|
| | MSE | MAE | P95 | MSE | MAE | P95 | MSE | MAE | P95 |
| DLinear | 0.850 | 0.683 | 1.786 | 0.861 | 0.694 | 1.770 | 0.808 | 0.679 | 1.694 |
| NLinear | 0.852 | 0.684 | 1.787 | 0.862 | 0.695 | 1.770 | 0.808 | 0.679 | 1.694 |
| GRU | 0.741 | 0.614 | **1.740** | 0.703 | 0.607 | 1.673 | 0.668 | 0.598 | 1.594 |
| TCN | **0.734** | **0.611** | 1.745 | **0.703** | **0.606** | **1.666** | **0.649** | **0.584** | **1.580** |
| PatchTST | 0.754 | 0.619 | 1.753 | 0.718 | 0.610 | 1.670 | 0.656 | 0.585 | 1.582 |
| TimesNet | 0.764 | 0.625 | 1.761 | 0.724 | 0.607 | 1.715 | 0.666 | 0.586 | 1.601 |
| iTransformer | 0.765 | 0.623 | 1.784 | 0.719 | 0.610 | 1.707 | 0.662 | 0.588 | 1.606 |

Table 5: Forward-prediction performance on our synthetic flapping-wing dataset using model seed 42. Errors are computed in normalized target space. Bold and underlined entries denote the best and second-best values within each prediction horizon.

| Model | $H = 1$ | | | $H = 16$ | | | $H = 32$ | | |
|---|---|---|---|---|---|---|---|---|---|
| | MSE | MAE | P95 | MSE | MAE | P95 | MSE | MAE | P95 |
| DLinear | 0.381 | 0.476 | 1.225 | 0.401 | 0.497 | 1.233 | 0.490 | 0.545 | 1.418 |
| NLinear | 0.387 | 0.491 | 1.204 | 0.413 | 0.502 | 1.265 | 0.444 | 0.524 | 1.315 |
| GRU | 0.146 | 0.278 | 0.827 | **0.170** | **0.307** | **0.858** | **0.187** | **0.323** | **0.916** |
| TCN | **0.142** | **0.273** | **0.801** | 0.185 | 0.317 | 0.898 | 0.209 | 0.336 | 0.938 |
| PatchTST | 0.294 | 0.418 | 1.105 | 0.289 | 0.414 | 1.083 | 0.276 | 0.403 | 1.064 |
| TimesNet | 0.211 | 0.342 | 0.966 | 0.231 | 0.361 | 1.011 | 0.228 | 0.360 | 1.011 |
| iTransformer | 0.224 | 0.356 | 0.989 | 0.215 | 0.346 | 0.968 | 0.217 | 0.348 | 0.990 |

Table 6: Forward-prediction performance on the independent real-world experimental flapping-wing dataset using model seed 42. The task uses the same shared kinematic and aerodynamic channels as the synthetic benchmark. Errors are computed in normalized target space. Bold and underlined entries denote the best and second-best values within each prediction horizon.

### Interpretation and Scope of the External Validation

The cross-dataset comparison provides several forms of external validation for our synthetic dataset. First, recurrent and convolutional temporal models form the leading architecture group in both datasets for short-horizon prediction. Second, DLinear and NLinear consistently rank near the bottom, indicating that both datasets require more than a direct linear temporal mapping. Third, forward prediction exhibits strong and statistically significant ranking agreement at $H = 1$ and $H = 16$, despite the synthetic and real-world trajectories originating from different systems. The results also identify a clear remaining domain gap. Long-horizon inverse prediction exhibits weaker ranking agreement, and the strong synthetic performance of PatchTST does not transfer to the real-world benchmark. The discrepancy may reflect differences in trajectory regularity, measurement noise, inverse identifiability, operating conditions, or long-range temporal structure. The present comparison does not isolate which of these factors is dominant.

The experiment is not a same-platform comparison between simulated and measured trajectories. Our synthetic dataset and the real-world dataset do not contain matched inputs, matched physical trials, or measurements from the same wing geometry and operating conditions. Moreover, each dataset is normalized independently using statistics estimated from its own training episodes. Similarity in normalized errors should therefore not be interpreted as equality of physical prediction error.

The results do not establish direct simulator fidelity or zero-shot simulation-to-real transfer. Instead, the comparison demonstrates that our synthetic dataset reproduces several nontrivial architecture-level and task-level learning patterns observed in independent real-world experimental data, particularly for forward and short-horizon inverse prediction. At the same time, the long-horizon inverse discrepancy provides a quantitative characterization of the remaining simulation-to-experiment gap.

## Dataset Conversion and Released Representation

### Episode Conversion and Quality Control

The synchronized simulator records are processed on an episode-by-episode basis. An initial transient interval is removed before quality screening and benchmark construction. Episodes are excluded when they contain non-finite core variables, have insufficient retained duration, exceed the prescribed wing-speed threshold, or exceed the allowable aggregate-load threshold. Each source episode is associated with a quality record containing its source identifier, original and retained lengths, retained duration, validity status, filtering reason, maximum left- and right-wing sampling speeds, maximum absolute aggregate load, mean forward and vertical body velocities, and mean absolute angle of attack.

For compatibility with fixed-length sequence-learning pipelines, each compact episode is stored at a target length of 470 samples. For an episode with retained length $T_i$:

- if $T_i = 470$, the sequence is stored without modification;
- if $T_i > 470$, 470 indices are selected at uniformly distributed positions over the retained sequence; and

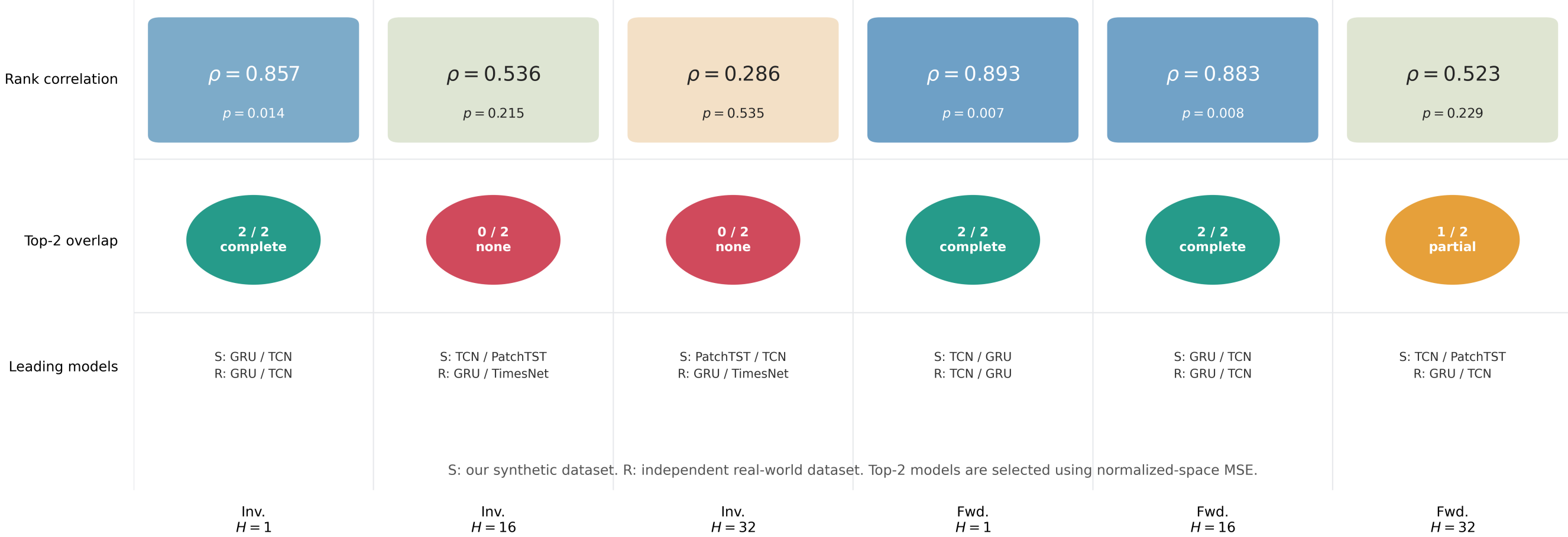

Figure 8: Agreement between our synthetic benchmark and the independent real-world experimental benchmark. The upper row reports the Spearman correlation between architecture rankings and its two-sided $p$-value. The middle row reports the overlap between the two best models in the two datasets. The lower row lists the two leading architectures on our synthetic data (S) and real-world data (R). Strong agreement is observed for inverse prediction at $H = 1$ and forward prediction at $H = 1$ and $H = 16$. The absence of Top-2 overlap for inverse prediction at $H = 16$ and $H = 32$ identifies the principal remaining cross-domain discrepancy.

| Task | $H$ | Best synthetic model | Best real-world model | Top-2 overlap | Spearman $\rho$ | $p$-value |
|---|---|---|---|---|---|---|
| Inverse | 1 | GRU | GRU | 2 / 2 | 0.857 | 0.014 |
| Inverse | 16 | TCN | GRU | 0 / 2 | 0.536 | 0.215 |
| Inverse | 32 | PatchTST | GRU | 0 / 2 | 0.286 | 0.535 |
| Forward | 1 | TCN | TCN | 2 / 2 | 0.893 | 0.007 |
| Forward | 16 | TCN / GRU | GRU | 2 / 2 | 0.883 | 0.008 |
| Forward | 32 | TCN | GRU | 1 / 2 | 0.523 | 0.229 |

Table 7: Agreement between MSE rankings on our synthetic dataset and the independent real-world experimental dataset. Spearman correlations are computed across the seven model families evaluated in both datasets. Top-2 overlap reports the number of shared architectures among the two best MSE results in each dataset. Ties in the available rounded synthetic results are assigned average ranks.

- if $T_i < 470$, the complete physical prefix is retained and the final valid observation is repeated to fill the remaining storage positions.

For padded episodes, a binary validity mask is defined as

$$m_t^{(i)} = \begin{cases} 1, & 1 \le t \le T_i, \\ 0, & T_i < t \le 470. \end{cases} \tag{28}$$

Episodes resampled to 470 positions have an all-valid mask. The mask is used during normalization and sample construction so that padded storage positions do not contribute to the benchmark statistics or sliding-window samples.

## Compatibility Storage and Channel Selection

For compatibility with the reference data pipelines, each fixed-length file stores a three-channel kinematic array, a five-channel response-side array, and a binary validity mask. Under the conversion settings used for the benchmark, the stored kinematic vector is

$$\widetilde{\mathbf{K}}(t) = [\phi_L(t) \quad 0 \quad \theta_L(t)]^\mathsf{T}, \tag{29}$$

where $\phi_L$ and $\theta_L$ denote the realized left-wing flap and twist angles, respectively. The central zero-valued channel is retained solely for compatibility with the reference storage format. The stored response-side vector is

$$\widetilde{\mathbf{R}}(t) = \begin{bmatrix} C_{F,L,y}(t) \\ C_{F,R,y}(t) \\ C_{F,\text{net},y}(t) \\ u_{\text{roll}}(t) \\ V_{\text{body},y}(t) \end{bmatrix}. \tag{30}$$

where $u_{\text{roll}}$ denotes the normalized roll-command channel and $V_{\text{body},y}$ denotes body vertical velocity.

The benchmark algorithms select kinematic channels 0 and 2 and response-side channels 0, 1, 2, and 4. The resulting

**Architecture Ranking Transfer from Our Synthetic Dataset to Real-World Experimental Data**

**Inverse prediction, $H = 1$** — $\rho = 0.857$, $p = 0.014$, **Top-2 overlap 2/2**

**Inverse prediction, $H = 16$** — $\rho = 0.536$, $p = 0.215$, **Top-2 overlap 0/2**

**Inverse prediction, $H = 32$** — $\rho = 0.286$, $p = 0.535$, **Top-2 overlap 0/2**

**Forward prediction, $H = 1$** — $\rho = 0.893$, $p = 0.007$, **Top-2 overlap 2/2**

**Forward prediction, $H = 16$** — $\rho = 0.883$, $p = 0.008$, **Top-2 overlap 2/2**

**Forward prediction, $H = 32$** — $\rho = 0.523$, $p = 0.229$, **Top-2 overlap 1/2**

Circles denote synthetic-data rankings; squares denote real-world rankings. Rank 1 corresponds to the lowest MSE. Synthetic metrics are available to three decimal places.

Figure 9: Architecture-ranking transfer from our synthetic dataset to the independent real-world experimental dataset. Circles denote rankings on our synthetic data, and squares denote rankings on real-world data. Rank 1 corresponds to the lowest normalized-space MSE. Each panel reports the Spearman rank correlation, its two-sided $p$-value, and the overlap between the two highest-ranked architectures. Ties in the available rounded synthetic results are assigned average ranks.

task-level representations are

$$\mathbf{K}(t) = [\phi_L(t) \quad \theta_L(t)]^{\mathsf{T}},$$

$$\mathbf{R}(t) = [C_{F,L,y}(t) \quad C_{F,R,y}(t) \quad C_{F,\mathrm{net},y}(t) \quad V_{\mathrm{body},y}(t)]^{\mathsf{T}}. \tag{31}$$

The zero-valued compatibility channel and normalized roll-command channel do not contribute to model inputs, prediction targets, training losses, or evaluation metrics.

## Episode Splits, Window Counts, and Normalization

A fixed episode-level manifest defines 1,600 training, 200 validation, and 200 test episodes. The manifest is validated for duplicate, overlapping, out-of-range, and missing indices and is reused across both task directions, all prediction horizons, and all evaluated architectures. Every window derived from a given episode remains in the same partition.

For an episode with valid length $T_i$, input-history length $L$, prediction horizon $H$, and stride $d$, the number of admissible windows is

$$N_i(L, H, d) = \max\left(0, \left\lfloor \frac{T_i - L - H}{d} \right\rfloor + 1\right). \tag{32}$$

The reported experiments use $L = 256$ and $d = 1$. The total number of samples in partition $\mathcal{P}$ is therefore

$$N_{\mathcal{P}}(L, H) = \sum_{i \in \mathcal{P}} N_i(L, H, 1). \tag{33}$$

Because admissibility is evaluated separately for every episode, no window crosses an episode boundary or contains a padded storage position.

| Horizon $H$ | Training | Validation | Test |
|---|---|---|---|
| 1 | 180,740 | 21,745 | 20,627 |
| 16 | 159,198 | 19,026 | 18,037 |
| 32 | 136,774 | 16,264 | 15,365 |

Table 8: Numbers of sliding-window samples in each episode-level partition. The forward and inverse tasks have identical sample counts.

Each selected input and target channel is standardized independently. The corresponding means and standard deviations are estimated using mask-valid observations from the training episodes only. Padded positions and all validation and test observations are excluded when fitting the standardizers. The resulting training-derived transformations are applied without modification during validation and testing. The fitted statistics are distributed with the released machine-readable benchmark metadata.

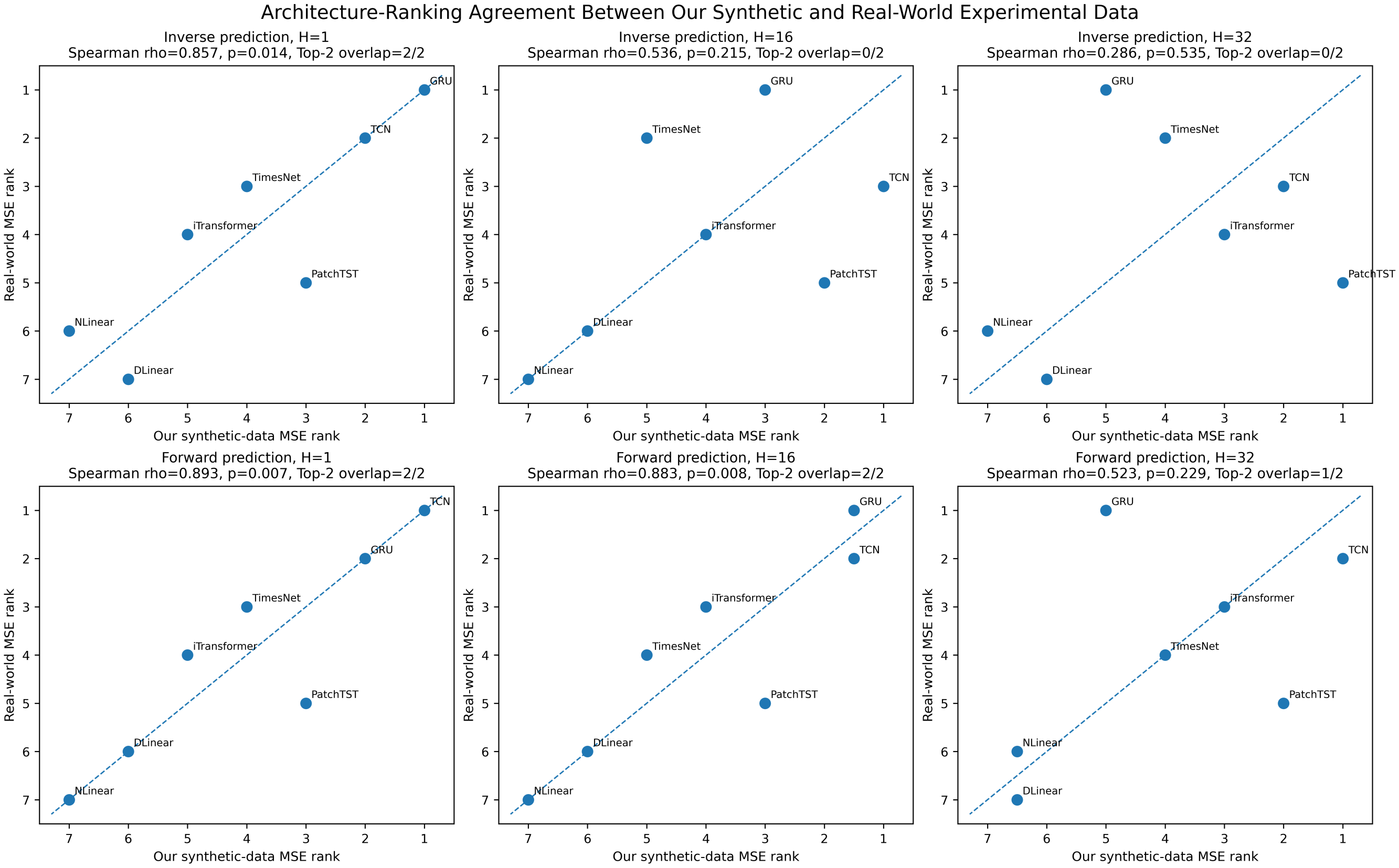

Figure 10: Paired architecture rankings on our synthetic dataset and the independent real-world experimental dataset. Each point represents one of the seven model families evaluated in both datasets. The dashed diagonal denotes identical synthetic and real-world MSE rankings, while displacement from the diagonal indicates a change in relative model performance. Each panel reports the Spearman rank correlation, its two-sided $p$-value, and the Top-2 architecture overlap. Agreement is strongest for inverse prediction at $H = 1$ and forward prediction at $H = 1$ and $H = 16$. Ties in the available rounded synthetic results are assigned average ranks.

## Inference Efficiency

All inference measurements were performed in float32 precision on a single AMD Instinct MI250X GCD using PyTorch 2.10.0 with ROCm 7.0. Each configuration was evaluated using 30 warm-up iterations followed by 200 synchronized forward passes at a batch size of one. Model loading, data loading, normalization, and disk access were excluded from the timed interval. The reported latency therefore represents single-sequence model-forward execution rather than end-to-end system latency. Although TimesNet was trained using single-node distributed data parallelism across eight GCDs, inference used one checkpoint replica on a single GCD. The reported TimesNet latency excludes distributed communication and is therefore directly comparable with the single-device inference latency of the remaining architectures.

Tables 9 and 10 report the median and 95th-percentile inference latencies, peak accelerator memory, and total parameter count for the forward and inverse tasks, respectively.

The efficiency results show that model size alone does not determine accelerator latency or predictive accuracy. NLinear and DLinear achieve the lowest single-sequence latencies, with median values of approximately $0.21$ ms and $0.43$ ms, respectively, but exhibit substantially larger prediction errors, particularly in the inverse task. TCN provides a strong balance between accuracy and computational cost, with a median latency of approximately $1.38$ ms while attaining the lowest forward MSE at all three horizons and ranking first or second by inverse MSE. PatchTST requires approximately $2.3$–$2.4$ ms per sequence and achieves the lowest long-horizon inverse error. GRU has substantially fewer parameters than PatchTST or iTransformer but requires approximately $9.0$ ms per forward pass. This latency is consistent with the sequential recurrent computation of GRU, which provides less temporal parallelism than the convolutional and attention-based alternatives. GRU nevertheless achieves the lowest one-step inverse error. TimesNet requires approx-imately 5.7–6.1 ms for single-sequence inference and has the largest total parameter count and peak memory usage among the evaluated models. Autoformer exhibits the highest single-sequence latency, approximately 12.2–12.5 ms, without a corresponding predictive-accuracy advantage.

The reported measurements represent model-forward execution rather than complete closed-loop control latency. A deployed system would additionally include sensor acquisition, normalization, communication, command generation, scheduling, and actuator delays.

## Kinematic and Multivariate Consistency

### Bilateral Kinematic Consistency and Command Tracking

Under the simulator coordinate convention, the left- and right-wing flap angles are mirror aligned, whereas the two twist angles follow the same sign convention:

$$\phi_L(t) \approx -\phi_R(t), \qquad \theta_L(t) \approx \theta_R(t). \tag{34}$$

The corresponding instantaneous residuals are defined as

$$r_\phi(t) = \left|\phi_L(t) + \phi_R(t)\right|, \qquad r_\theta(t) = \left|\theta_L(t) - \theta_R(t)\right|. \tag{35}$$

The residual distributions are summarized using their mean, root-mean-square value, upper quantiles, and maximum.

Figure 5 evaluates the bilateral coordinate convention and flap-command tracking. After sign alignment, the left- and right-wing flap trajectories exhibit close agreement over the flap cycle, while the twist trajectories show similarly small discrepancies under their shared sign convention. These results support the use of the realized left-wing flap and twist angles as representative benchmark variables.

The commanded-versus-realized distribution is concentrated near the identity relation, indicating close flap-command tracking. The flap–twist phase portrait additionally shows the coupled cyclic evolution of the realized kinematic variables and their variation across episodes.

## Model and Training Configurations

All configurations reported in this section are extracted from the saved metadata of the final benchmark runs. A value is listed as an architecture-specific hyperparameter only when it is consumed by the corresponding model implementation. Unused generic parser defaults are omitted.

All models follow the optimization and model-selection protocol summarized in Table 11. Architecture-specific settings are reported in Table 12, and the corresponding execution configurations are reported in Table 13.

## Metric Implementation Details

Aggregate MSE and MAE are computed over all scalar target values in the held-out test partition. Squared and absolute errors are accumulated across test batches, prediction steps, and target channels before the final aggregate metrics are calculated. For $E_{95}$, absolute errors are pooled over the same dimensions before the empirical 95th percentile is evaluated. Channel-wise MSE and MAE are computed separately for each predicted variable, whereas the reported aggregate $E_{95}$ is not channel specific.

For the forward task, the channel-wise metrics characterize the left-wing, right-wing, and aggregate vertical force coefficients and the body vertical velocity. For the inverse task, they characterize the realized flap- and twist-angle predictions. All predictive metrics are evaluated in normalized target space. Forward and inverse results are evaluated independently because the two tasks have different target variables, output dimensions, and target standardizers.

## Additional Benchmark Visualizations

Figure 11 visualizes the MSE-based model rankings across prediction horizons. The forward ranking is stable, with TCN remaining first at all three horizons. By contrast, the inverse winner changes from GRU at $H = 1$ to TCN at $H = 16$ and PatchTST at $H = 32$. The figure also highlights the deterioration of GRU as the inverse prediction horizon increases.

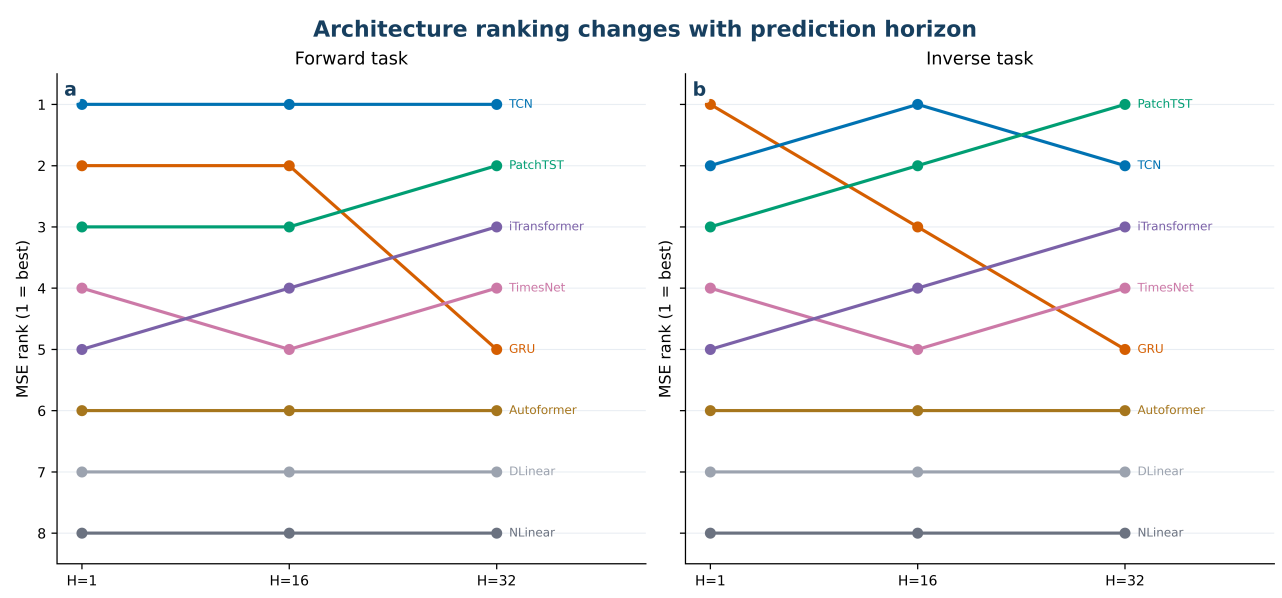

Figure 11: Horizon-dependent architecture rankings based on normalized test MSE. (a) Forward prediction, for which TCN remains the highest-ranked model across all three horizons. (b) Inverse prediction, for which the leading model changes from GRU at $H = 1$ to TCN at $H = 16$ and PatchTST at $H = 32$. Lower ranks indicate better performance. Rankings are determined using the unrounded MSE values.

## Inference Efficiency

All inference measurements were performed in float32 precision on a single AMD Instinct MI250X GCD using PyTorch 2.10.0 with ROCm 7.0. Each configuration was evaluated using 30 warm-up iterations followed by 200 synchronized forward passes at a batch size of one. Model loading, data loading, normalization, and disk access were excluded from the timed interval. The reported latency therefore represents single-sequence model-forward execution rather than end-to-end system latency. Although TimesNet was trained using single-node distributed data parallelism across eight GCDs, inference used one checkpoint replica on a single GCD. The reported TimesNet latency excludes distributed communication and is therefore directly comparable with the single-device inference latency of the remaining architectures. Tables 9 and 10 report the median and 95th-percentile inference latencies, peak accelerator memory, and total parameter count for the forward and inverse tasks, respectively.

| Model | $H = 1$ | | | | $H = 16$ | | | | $H = 32$ | | | |
|---|---|---|---|---|---|---|---|---|---|---|---|---|
| | Med. (ms) | P95 (ms) | Mem. (MiB) | Total Params. | Med. (ms) | P95 (ms) | Mem. (MiB) | Total Params. | Med. (ms) | P95 (ms) | Mem. (MiB) | Total Params. |
| NLinear | **0.212** | **0.230** | 76.0 | 269 | **0.211** | **0.229** | 76.0 | 4,124 | **0.211** | **0.215** | 76.0 | 8,236 |
| DLinear | 0.434 | 0.445 | 76.0 | 526 | 0.432 | 0.449 | 76.0 | 8,236 | 0.436 | 0.448 | 76.1 | 16,460 |
| TCN | 1.380 | 1.403 | 76.7 | 87,364 | 1.381 | 1.406 | 76.7 | 91,264 | 1.378 | 1.400 | 76.7 | 95,424 |
| iTransformer | 2.130 | 2.167 | 77.2 | 298,253 | 2.115 | 2.156 | 77.3 | 300,188 | 2.119 | 2.159 | 77.2 | 302,252 |
| PatchTST | 2.346 | 2.384 | 79.8 | 271,373 | 2.413 | 2.526 | 80.1 | 332,828 | 2.323 | 2.366 | 80.3 | 398,380 |
| TimesNet | 6.071 | 6.129 | 106.0 | 4,666,895 | 5.803 | 5.872 | 104.7 | 4,670,750 | 5.811 | 5.877 | 105.4 | 4,674,862 |
| GRU | 9.048 | 9.262 | 76.4 | 13,316 | 8.954 | 9.120 | 76.4 | 17,216 | 9.068 | 9.273 | 76.5 | 21,376 |
| Autoformer | 12.358 | 12.434 | 85.5 | 483,342 | 12.286 | 12.346 | 85.5 | 483,342 | 12.467 | 12.584 | 85.5 | 483,342 |

Table 9: Batch-one inference efficiency for the forward task on one AMD Instinct MI250X GCD. Med. and P95 denote the median and 95th-percentile forward-pass latencies, respectively; Mem. denotes peak accelerator memory, and Total Params. denotes the total parameter count. Bold and underlined latency values identify the best and second-best results within each prediction horizon.

| Model | $H = 1$ | | | | $H = 16$ | | | | $H = 32$ | | | |
|---|---|---|---|---|---|---|---|---|---|---|---|---|
| | Med. (ms) | P95 (ms) | Mem. (MiB) | Total Params. | Med. (ms) | P95 (ms) | Mem. (MiB) | Total Params. | Med. (ms) | P95 (ms) | Mem. (MiB) | Total Params. |
| NLinear | **0.213** | **0.234** | 76.0 | 267 | **0.210** | **0.231** | 76.0 | 4,122 | **0.211** | **0.230** | 76.0 | 8,234 |
| DLinear | 0.432 | 0.448 | 76.0 | 524 | 0.437 | 0.447 | 76.0 | 8,234 | 0.436 | 0.449 | 76.1 | 16,458 |
| TCN | 1.380 | 1.386 | 76.7 | 87,746 | 1.381 | 1.407 | 76.7 | 89,696 | 1.383 | 1.408 | 76.7 | 91,776 |
| iTransformer | 2.112 | 2.134 | 77.2 | 298,251 | 2.111 | 2.143 | 77.2 | 300,186 | 2.118 | 2.160 | 77.3 | 302,250 |
| PatchTST | 2.335 | 2.371 | 80.2 | 271,371 | 2.343 | 2.375 | 80.4 | 332,826 | 2.346 | 2.381 | 80.7 | 398,378 |
| TimesNet | 5.994 | 6.264 | 106.6 | 4,667,919 | 5.887 | 5.963 | 104.9 | 4,671,774 | 5.721 | 5.833 | 106.5 | 4,675,886 |
| GRU | 9.089 | 9.283 | 76.4 | 13,570 | 9.087 | 9.279 | 76.4 | 15,520 | 9.061 | 9.249 | 76.4 | 17,600 |
| Autoformer | 12.357 | 13.539 | 85.5 | 485,902 | 12.214 | 12.383 | 85.5 | 485,902 | 12.423 | 12.591 | 85.5 | 485,902 |

Table 10: Batch-one inference efficiency for the inverse task on one AMD Instinct MI250X GCD. Med. and P95 denote the median and 95th-percentile forward-pass latencies, respectively; Mem. denotes peak accelerator memory, and Total Params. denotes the total parameter count. Bold and underlined latency values identify the best and second-best results within each prediction horizon.

The efficiency results show that model size alone does not determine accelerator latency or predictive accuracy. NLinear and DLinear achieve the lowest single-sequence latency but exhibit substantially larger prediction errors, particularly in the inverse task. TCN provides a strong balance between accuracy and computational cost, with a median latency of approximately 1.43 ms while attaining the lowest forward MSE at all three horizons and ranking first or second by inverse MSE. PatchTST requires approximately 2.40 ms per sequence and achieves the lowest long-horizon inverse error. GRU has substantially fewer parameters than PatchTST or iTransformer but requires approximately 10.2 ms per forward pass. This latency is consistent with the sequential recurrent computation of GRU, which provides less temporal parallelism than the convolutional and attention-based alternatives. GRU nevertheless achieves the lowest one-step inverse error. TimesNet requires approximately 6.0–6.3 ms for single-sequence inference and has the largest total parameter count and peak memory usage among the evaluated models. Autoformer exhibits the highest single-sequence latency, approximately 13 ms, without a corresponding predictive-accuracy advantage. The reported measurements represent model-forward execution rather than complete closed-loop control latency. A deployed system would additionally include sensor acquisition, normalization, communication, command generation, scheduling, and actuator delays.

| Setting | Value |
|---|---|
| Input-history length $L$ | 256 |
| Prediction horizons $H$ | $\{1, 16, 32\}$ |
| Window stride | 1 |
| Maximum epochs | 500 |
| Optimizer | AdamW |
| Initial learning rate | $10^{-3}$ |
| Weight decay | $10^{-4}$ |
| Gradient-norm limit | 1.0 |
| LR reduction factor | 0.5 |
| LR-scheduler patience | 10 epochs |
| Minimum learning rate | $10^{-6}$ |
| Early-stopping patience | 50 epochs |
| Model seed | 42 |
| Training objective | Normalized-space MSE |
| Checkpoint criterion | Minimum validation MSE |

Table 11: Shared optimization and model-selection settings used throughout the benchmark.

| Model | Architecture-specific configuration |
|---|---|
| DLinear | Moving-average length 25; shared linear projection (`individual=False`) |
| NLinear | Last-value normalization; shared linear projection (`individual=False`) |
| GRU | Hidden dimension 64; one recurrent layer; dropout 0.05 |
| TCN | Four convolutional stages with channels $[64, 64, 64, 64]$; kernel size 3; exponentially increasing dilation; dropout 0.10 |
| PatchTST | $d_{\text{model}} = 128$; 8 attention heads; 2 encoder layers; $d_{\text{ff}} = 256$; patch length 16; patch stride 8; dropout 0.10 |
| TimesNet | $d_{\text{model}} = 128$; 2 encoder layers; $d_{\text{ff}} = 256$; top-$k = 3$; 3 convolutional kernels; dropout 0.10 |
| iTransformer | $d_{\text{model}} = 128$; 8 attention heads; 2 encoder layers; $d_{\text{ff}} = 256$; dropout 0.10 |
| Autoformer | $d_{\text{model}} = 128$; 8 attention heads; 2 encoder layers; 1 decoder layer; $d_{\text{ff}} = 256$; factor 3; moving-average length 25; dropout 0.10 |

Table 12: Architecture-specific settings used in the final benchmark runs. Task- and horizon-specific output mappings are included in each trained configuration.

| Models | Devices | batch | Global | Drop |
|---|---|---|---|---|
| DLinear, NLinear, GRU | 1 | 512 | 512 | No |
| TCN, PatchTST, iTransformer | 1 | 512 | 512 | No |
| Autoformer | 1 | 512 | 512 | No |
| TimesNet | 8 | 128 | 1,024 | Yes |

Table 13: Training execution configurations. TimesNet was trained using single-node distributed data parallelism across eight accelerator processes; the remaining architectures used single-device training.